\documentclass[sigconf]{acmart}

\usepackage{algorithm}
\usepackage{algorithmic, eucal}
\usepackage{graphicx}
\usepackage{url}
\usepackage{soul}
\usepackage{amsmath}
\usepackage{balance}
\usepackage{booktabs}

\newcommand{\eg}{\textit{e.g.}}
\newcommand{\ie}{\textit{i.e.}}
\DeclareMathOperator*{\argmin}{arg\,min}

\newcommand{\xvec}{{\mathbf{x}}}
\newcommand{\yij}{y_{\{\mathbf{x}_i,\mathbf{x}_j\}}}

\newcommand{\uvec}{{\mathbf{u}}}
\newcommand{\avec}{{\mathbf{a}}}

\AtBeginDocument{%
  \providecommand\BibTeX{{%
    \normalfont B\kern-0.5em{\scshape i\kern-0.25em b}\kern-0.8em\TeX}}}

\copyrightyear{2023}
\acmYear{2023}
\setcopyright{acmlicensed}
\acmConference[KDD '23] {Proceedings of the 29th ACM SIGKDD Conference on Knowledge Discovery and Data Mining}{August 6--10, 2023}{Long Beach, CA, USA.}
\acmBooktitle{Proceedings of the 29th ACM SIGKDD Conference on Knowledge Discovery and Data Mining (KDD '23), August 6--10, 2023, Long Beach, CA, USA}
\acmPrice{15.00}
\acmISBN{979-8-4007-0103-0/23/08}
\acmDOI{10.1145/3580305.3599302}

\begin{document}

\title[Deep Weakly-supervised Anomaly Detection]{Deep Weakly-supervised Anomaly Detection}


\author{Guansong Pang}
\authornote{Corresponding author: G. Pang}
\affiliation{%
  \institution{Singapore Management University}
  \city{Singapore}
  \country{Singapore}
  \postcode{178902}
}
\email{pangguansong@gmail.com}

\author{Chunhua Shen}
\affiliation{%
  \institution{Zhejiang University}
 \city{Hangzhou}
 \country{China}
  \postcode{310027}
}
\email{chhshen@gmail.com}

\author{Huidong Jin}
\affiliation{%
  \institution{Data61}
  \city{Canberra}
  \country{Australia}
  \postcode{2601}
}
\email{warren.jin@csiro.au}

\author{Anton van den Hengel}
\affiliation{%
  \institution{University of Adelaide}
  \city{Adelaide}
  \country{Australia}
  \postcode{5005}
}
\email{anton.vandenhengel@adelaide.edu.au}

\renewcommand{\shortauthors}{Pang, et al.}

\begin{abstract}
Recent semi-supervised anomaly detection methods that are trained using small labeled anomaly examples and large unlabeled data (mostly normal data) have shown largely improved performance over unsupervised methods. However, 
these methods often focus on fitting abnormalities illustrated by the given anomaly examples only (\ie, seen anomalies), and consequently they fail to generalize to those that are not, \ie, new types/classes of anomaly unseen during training.
To detect both seen and unseen anomalies, we introduce a novel deep weakly-supervised approach, namely \underline{P}airwise \underline{Re}lation prediction \underline{Net}work (PReNet), that \textit{learns pairwise relation features and anomaly scores} by predicting the relation of any two randomly sampled training instances, in which the pairwise relation can be anomaly-anomaly, anomaly-unlabeled, or unlabeled-unlabeled. Since unlabeled instances are mostly normal, the relation prediction enforces a joint learning of anomaly-anomaly, anomaly-normal, and normal-normal pairwise discriminative patterns, respectively. PReNet can then detect any seen/unseen abnormalities that fit the learned pairwise abnormal patterns, or deviate from the normal patterns. Further, this pairwise approach also seamlessly and significantly augments the training anomaly data.
Empirical results on 12 real-world datasets show that PReNet significantly outperforms nine competing methods in detecting seen and unseen anomalies. We also theoretically and empirically justify the robustness of our model w.r.t. anomaly contamination in the unlabeled data. The code is available at \renewcommand\UrlFont{\color{blue}\tt}\url{https://github.com/mala-lab/PReNet}.
\end{abstract}

%


\maketitle

\section{Introduction}\label{sec:introduction}

Anomaly detection (AD) aims at identifying exceptional data instances that deviate significantly from the majority of data. It is of critical practical importance due to its broad applications
in defensing against cyber-crimes (\eg, network intrusions), user misbehavior (\eg, fraudulent user accounts/reviews), web advertising abuses, and adverse drug reactions, to name a few \cite{aggarwal2017outlieranalysis,jin2008mining,jin2010signaling}. Numerous AD methods have been introduced, most of which are unsupervised methods working on entirely unlabeled (mostly normal data) data \cite{aggarwal2017outlieranalysis,pang2021survey}. The popularity of the unsupervised methods is mainly because they avoid the significant cost of manually labeling large-scale anomaly data, which is required to support fully-supervised approaches. However, they operate without knowing what true anomalies look like, and as a result, they identify many noisy or uninteresting isolated data instances as anomalies, leading to high detection errors. 

Recent semi-supervised AD methods \cite{pang2018repen,zhang2018anomaly,pang2019devnet,ruff2019deep,ding2021few,pang2021explainable} aim to bridge the gap between supervised and unsupervised AD by utilizing a limited number of anomaly examples
to train anomaly-informed detection models. This line of research is motivated by the fact that a small set of labeled anomaly examples (\eg, some successfully detected anomalous events) can often be made available with a small cost in real-world applications. These labeled anomalies provide a strong indication of the anomalies of interest and can substantially enhance the detection accuracy \cite{pang2018repen,zhang2018anomaly,pang2019devnet,ruff2019deep,ding2021few,pang2021explainable}. 
However, anomalies are unknown abnormal events, so the labeled anomaly examples typically provides only an \textit{incomplete illustration of anomalies}. The current methods focus on fitting the abnormalities illustrated by the small anomaly examples (\ie, \textbf{seen anomalies}); they fail to generalize to those that are not, \ie, new types/classes of anomaly unseen during training (\textbf{unseen anomalies}), such as zero-day attacks \cite{wang2013k,pang2021toward} and novel defects/planet surfaces \cite{kerner2020comparison,pang2021explainable,ding2022catching}. Also, their performance in detecting seen anomalies is restricted due to the lack of large training anomaly data.

To tackle these issues, this work considers the problem of \textbf{weakly-supervised AD}, or alternatively open-set supervised AD, that aims to detect both seen and unseen anomalies given an incomplete illustration of anomaly classes seen during training. To this end,
we introduce a novel deep AD approach, namely \underline{P}airwise \underline{Re}lation prediction \underline{Net}work (\textbf{PReNet}), that learns pairwise relation features and anomaly scores by predicting the relation of any two training instances randomly sampled from the small anomaly data and the unlabeled data, in which the pairwise relation labels can be \textit{anomaly-anomaly, anomaly-unlabeled, or unlabeled-unlabeled}. During inference, a test instance is considered as an anomaly, if it fits well to the first two types of pairs, or deviates from the last pair type, when paired with a random training instance. In essence, our approach unifies the relation prediction and anomaly scoring, and learns to assign larger prediction scores (\ie, anomaly scores) to the instance pairs that contain anomalies than the other instance pairs.

Our key insight is that since the unlabeled data is often mostly normal, the pairwise class labels offer rich three-way pairwise relation information that supports a joint learning of diverse discriminative patterns, including \textit{anomaly-anomaly, anomaly-normal, and normal-normal pairwise feature patterns}, avoiding the fitting of the seen abnormalities only. Our approach can then detect any seen/unseen abnormalities that fit the learned pairwise abnormal patterns, or deviate from the normal patterns.
Further, the pairwise relation formulation seamlessly generates large-scale anomaly-informed surrogate class labels, \ie, the anomaly-anomaly and anomaly-unlabeled labels vs unlabeled-unlabeled. This significantly extends the training anomaly data, supporting effective training of a generalized detection model with the limited labeled data.    

In summary, this work makes four main contributions:
\begin{itemize}
    \item \textbf{Problem and Approach}. We consider the under-explored yet crucial problem -- weakly-supervised anomaly detection -- and propose a novel pairwise relation learning approach PReNet to address the problem. PReNet learns diverse discriminative pairwise relation features, offering more generalized detection models than existing methods.
    
    \item \textbf{Detection Model}. PReNet is instantiated to a novel detection model that
    learns pairwise anomaly scores by minimizing a three-way prediction loss using a relation neural network. The model is trained with the support of significantly augmented pairwise anomaly data, enabling effective training of a generalized detection model.
    
    \item \textbf{Robustness}. We theoretically and empirically show that PReNet can effectively leverage the large unlabeled data while being tolerant to anomaly contamination.

    \item \textbf{Large Empirical Support}. Our empirical results on 12 real-world datasets show that PReNet (i) significantly outperforms nine state-of-the-art (SOTA) competing methods in detecting seen and unseen anomalies,
    and (ii) 
    obtains substantially better sample efficiency, \eg, it 
    requires 50\%-87.5\% less labeled anomaly data to perform comparably well to, or better than, the best competing models.

\end{itemize}

\vspace{-0.1cm}

\section{Related Work}

\noindent\textbf{Toward Supervised Anomaly Detection.}
Previous semi-supervised AD methods \cite{noto2012frac,gornitz2013semisupervisedad,ienco2017semisupervised,chandola2009anomaly,sperl2020activation} focus on leveraging labeled normal instances to learn patterns of the normal class. 
Since a small amount of anomaly data is often available in many real-world applications, recent semi-supervised methods \cite{mcglohon2009snare,tamersoy2014guilt,zhang2018anomaly,pang2018repen,ding2021few,ruff2019deep,zhao2018xgbod} are dedicated to utilizing small labeled anomaly data to learn anomaly detectors, \eg, label propagation \cite{mcglohon2009snare,tamersoy2014guilt,vercruyssen2018semi}, representation learning \cite{pang2018repen,ruff2019deep,zhou2022unseen}, classification models \cite{zhang2018anomaly}, or newly proposed loss functions \cite{pang2019devnet,ding2021few}, and they show that these limited labeled anomalies can substantially improve the detection accuracy. 
Among them, DevNet \cite{pang2019devnet,pang2021explainable} and Deep SAD (DSAD) \cite{ruff2019deep,liznerski2020explainable} are two most relevant methods, which achieve impressive detection performance
by fitting a Gaussian prior-driven anomaly score distribution and a one-class hypersphere, respectively.
However, they are prone to overfitting the given anomaly examples due to the lack of proper regularization and large-scale, diversified training anomaly samples. To address this issue, weakly-supervised AD \cite{pang2021explainable,pang2021toward,zhou2022feature,jiang2023weakly} and open-set supervised AD \cite{ding2022catching} tasks are recently introduced, aiming to detect both seen and unseen anomalies. We follow this line and introduce a novel pairwise relation learning approach.

This research line is also relevant to few-shot learning \cite{fei2006oneshot,vinyals2016matching,snell2017protonet,wang2020generalizing,sung2018learning,lai2021learning} and positive and unlabeled data (PU) learning \cite{li2003pul,elkan2008pul,sansone2018pul,bekker2020learning,ju2020pumad,chen2022weakly,luo2018pu} due to the availability of the limited labeled positive instances (anomalies), but they are very different in that these two areas assume that the few labeled instances share the same intrinsic class structure as the other instances within the same class (\ie, the anomaly class), whereas the seen anomalies and the unseen anomalies may have completely different class structures.

\noindent\textbf{Deep Anomaly Detection.}
Traditional AD approaches are often ineffective in high-dimensional or non-linear separable data due to the curse of dimensionality and the deficiency in capturing the non-linear relations \cite{aggarwal2017outlieranalysis,pang2021survey,zhang2009new}. Deep AD has shown promising results in handling those complex data, of which most methods are based on pre-trained features \cite{reiss2021panda,roth2022towards}, or features learned by using autoencoder- \cite{chen2017autoencoder,gong2019memorizing,zhang2019deep,park2020learning} or generative adversarial network- \cite{schlegl2017gan,zenati2018gan,li2019mad,zhang2022deep} based objectives. One issue with these methods is that these feature representations are not primarily optimized to detect anomalies. Some very recent methods \cite{zong2018autoencoder,pang2018repen,ruff2018deepsvdd,zheng2019oneclass} address this issue by learning representations tailored for specific anomaly measures, \eg, cluster membership-based measure in \cite{zong2018autoencoder,song2021deep}, distance-based measure in \cite{pang2018repen,wang2020unsupervised} and one-class classification-based measure in \cite{ruff2018deepsvdd,zheng2019oneclass,ruff2019deep,goyal2020drocc,sohn2020learning,chen2022deep}.
However, they still focus on optimizing the feature representations. By contrast, our model unifies representation learning and anomaly scoring into one pipeline to directly optimize anomaly scores, yielding more optimized anomaly scores. Further, these methods overwhelmingly focus on unsupervised/semi-supervised settings where detection models are trained on unlabeled data or exclusively normal data, which fail to utilize the valuable labeled anomaly data as available in many real-world applications. Some recent studies such as DevNet \cite{pang2019devnet,pang2021explainable,ding2021few} directly optimize the anomaly scores via a loss function called deviation loss \cite{pang2019devnet,pang2021explainable}. Our method instead uses a novel formulation of pairwise relation prediction to achieve the goal, which shows to be significantly better than the deviation loss. 

Additionally, our relation prediction is formulated as a weakly-supervised three-way ordinal regression task which is different from \cite{pang2020self} in both of the targeted problem and the approach taken since \cite{pang2020self} uses self-trained ordinal regression for unsupervised AD. Also, anomaly contamination estimation \cite{perini2020class,perini2022estimating,perini2022transferring} can help estimate the proportion of anomalies in the unlabeled data, which may be used as a prior for empowering weakly-supervised AD. 

\vspace{-0.05cm}    
\section{The Proposed Approach}

\subsection{Overview of Our Approach PReNet}
\subsubsection{Problem Statement} Given a training dataset $\mathcal{X}=\{ \mathbf{x}_{1}, \mathbf{x}_{2}, \cdots,\\ \mathbf{x}_{N}, \mathbf{x}_{N+1}, \cdots, \mathbf{x}_{N+K} \}$,
with $\mathbf{x}_{i} \in \mathbb{R}^{D}$, where $\mathcal{U}=\{ \mathbf{x}_{1}, \mathbf{x}_{2}, \cdots, \mathbf{x}_{N}\}$ is a large unlabeled dataset and $\mathcal{A}=\{\mathbf{x}_{N+1}, \mathbf{x}_{N+2}, \cdots, \mathbf{x}_{N+K} \}$ ($K\ll N$) is a small set of labeled anomaly examples that often do not illustrate every possible class of anomaly, our goal is to learn a scoring function $\phi: \mathcal{X} \rightarrow \mathbb{R}$ that assigns anomaly scores to data instances in a way that we have $\phi(\mathbf{x}_{i}) > \phi(\mathbf{x}_{j})$ if $\mathbf{x}_{i}$ is an anomaly (despite it is a seen or unseen anomaly) and $\mathbf{x}_{j}$ is a normal instance.

\subsubsection{Anomaly-informed Pairwise Relation Prediction} In our proposed approach PReNet, we formulate the problem as a pairwise relation prediction-based anomaly score learning, in which we learn to discriminate three types of random instance pairs, including anomaly-anomaly pairs, anomaly-unlabeled pairs, unlabeled-unlabeled pairs. 
The formulation unifies the relation prediction and anomaly scoring, and helps enforce the model to assign substantially larger prediction scores (\ie, anomaly scores) to the instance pairs that contain anomalies than the other instance pairs. By doing so, the model learns diverse discriminative pairwise relation features embedded in the three-way pairwise interaction data. This way helps alleviate the overfitting of the seen anomalies as the model is regularized by simultaneously learning a variety of pairwise normality/abnormality patterns, rather than the seen abnormalities only. Further, the pairwise relation labels generate significantly more labeled training data than the original data, offering sufficiently large surrogate labeled data to train a generalized detection model.

\begin{figure}[h!]
  \centering
    \includegraphics[width=0.49\textwidth]{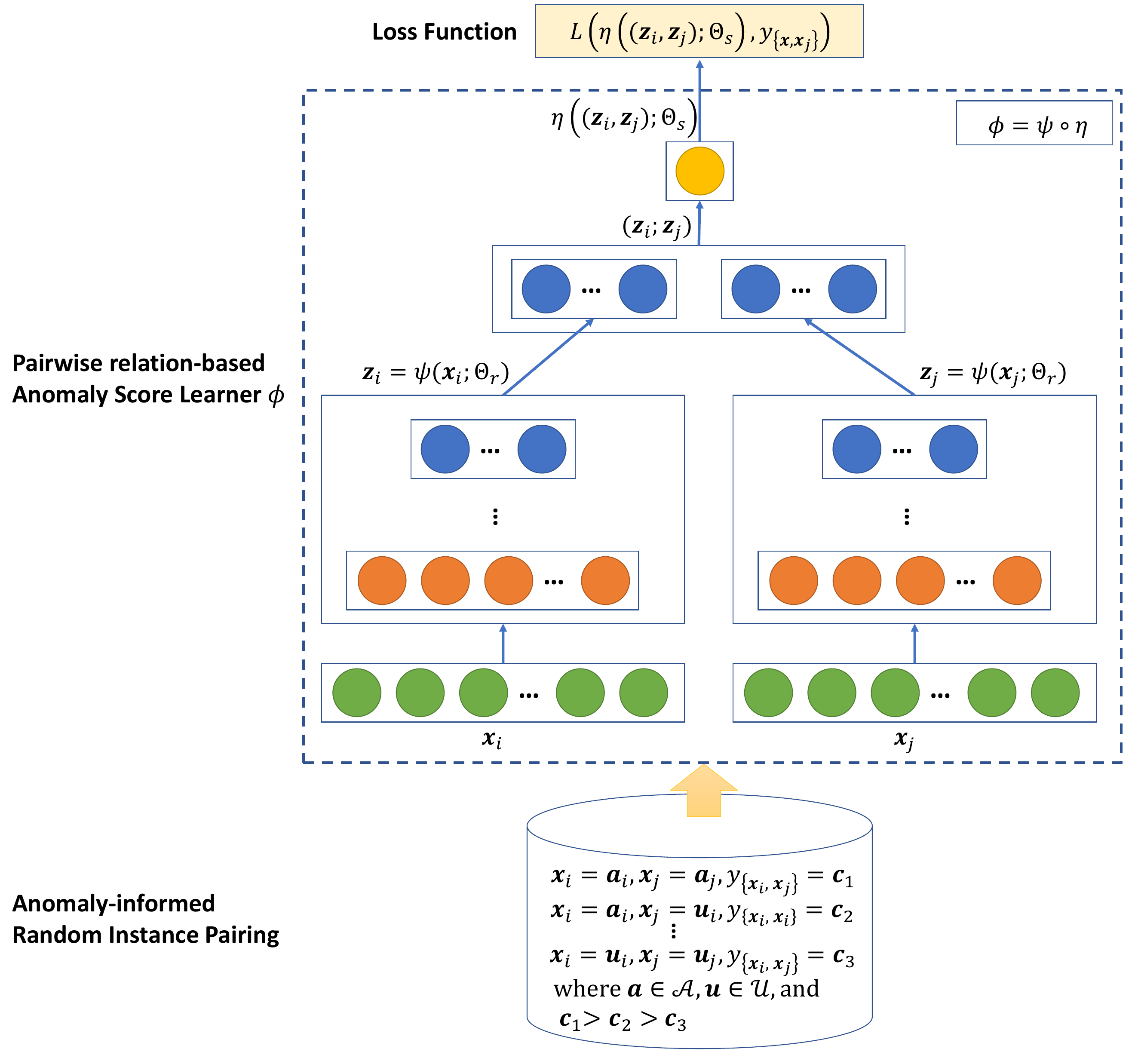}
  \caption{Overview of PReNet. It takes anomaly-anomaly, anomaly-unlabeled, and unlabeled-unlabeled instance pairs as input and learns pairwise anomaly scores by discriminating these three types of linear pairwise interactions.}
  \label{fig:framework}
\end{figure}

Specifically, as shown in Fig. \ref{fig:framework}, our approach consists of two main modules: \textit{anomaly-informed random instance pairing} and \textit{pairwise relation-based anomaly score learning}. The first module generates an instance pair dataset $\mathcal{P} = \big\{ \big(\mathbf{x}_{i}, \mathbf{x}_{j}, \yij \big) \,| \, \mathbf{x}_{i}, \mathbf{x}_{j} \in \mathcal{X} \; \text{and}\; \yij \in \mathbb{N} \big\}$, where each pair $\{\mathbf{x}_{i}, \mathbf{x}_{j}\}$ has one of the three pairwise relations: $C_{\mathbf{\{a,a\}}}$, $C_{\mathbf{\{a,u\}}}$ and $C_{\mathbf{\{u,u\}}}$ ($\mathbf{a} \in \mathcal{A}$ and $\mathbf{u} \in \mathcal{U}$) and $\mathbf{y} \in \mathbb{N}^{|\mathcal{P}|}$ is an ordinal class feature with \textit{decreasing} value assignments to the respective $C_{\mathbf{\{a,a\}}}$, $C_{\mathbf{\{a,u\}}}$ and $C_{\mathbf{\{u,u\}}}$ pairs, \ie, $y_{\mathbf{\{a,a\}}} > y_{\mathbf{\{a,u\}}} > y_{\mathbf{\{u,u\}}}$. These pairwise labels are set to be ordinal values to enable the anomaly score learning module $\phi: \mathcal{P} \rightarrow \mathbb{R}$, which can be treated as jointly learning a feature learner $\psi$ and a relation (anomaly score) learner $\eta$. $\phi$ is trained in an end-to-end manner to learn the pairwise anomaly scores using $\mathcal{P}$.

\subsection{The Instantiated Model}

The two modules of PReNet are specified as follows.

\subsubsection{Anomaly-informed Random Instance Pairing}

In this module, PReNet generates large-scale instance pairs with surrogate class labels to provide large labeled data for training subsequent pairwise relation prediction models. Specifically, instance pairs are created with instances randomly sampled from the small anomaly set $\mathcal{A}$ and the large unlabeled dataset $\mathcal{U}$. A pairwise class label is then assigned to each instance pair, such that $y_{\mathbf{\{a,a\}}}=c_1$, $y_{\mathbf{\{a,u\}}}=c_2$, $y_{\mathbf{\{u,u\}}}=c_3$ and $c_1>c_2>c_3\geq0$. By doing so, we efficiently synthesize $\mathcal{A}$ and $\mathcal{U}$ to produce a large labeled dataset $\mathcal{P} = \big\{ \big(\mathbf{x}_{i}, \mathbf{x}_{j}, \yij \big)  \,| \, \mathbf{x}_{i}, \mathbf{x}_{j} \in \mathcal{X} \; \text{and}\; \yij \in \mathbb{N} \big\}$. 

The resulting $\mathcal{P}$ contains critical information for discriminating anomalies from normal instances. This is because $y_{\mathbf{\{a,a\}}}$, $y_{\mathbf{\{a,u\}}}$ and $y_{\mathbf{\{u,u\}}}$ are approximately anomaly-anomaly, anomaly-normal and normal-normal pairs, respectively, as $\mathcal{U}$ is typically dominated by normal instances (per definition of anomaly \cite{aggarwal2017outlieranalysis,chandola2009anomaly}). A few $y_{\mathbf{\{a,u\}}}$ and $y_{\mathbf{\{u,u\}}}$ pairs may be noisy pairs due to anomaly contamination in $\mathcal{U}$, but we show that PReNet is robust to these noisy pairs

\subsubsection{Pairwise Relation-based Anomaly Score Learning}\label{subsubsec:scorelearning}

A pairwise anomaly score learner $\phi: \mathcal{P} \rightarrow \mathbb{R}$ is then introduced to take $\mathcal{P}$ as input to learn the anomaly scores of instance pairs. Let $\mathcal{Z} \in \mathbb{R}^{M}$ be an intermediate representation space, we define a two-stream anomaly scoring network $\phi\big((\cdot,\cdot); \Theta\big):\mathcal{P} \rightarrow \mathbb{R}$ as a sequential combination of a feature learner $\psi(\cdot; \Theta_{r}): \mathcal{X} \rightarrow \mathcal{Z}$ and an anomaly scoring function $\eta\big((\cdot,\cdot); \Theta_{s}\big): (\mathcal{Z},\mathcal{Z}) \rightarrow \mathbb{R}$, where $\Theta=\{\Theta_{r}, \Theta_{s}\}$. Specifically, $\psi(\cdot; \Theta_{r})$ is a neural \textit{feature learner} with $H \in \mathbb{N}$ hidden layers and the weight parameters $\Theta_{r}$.
\begin{equation}
    \mathbf{z} = \psi(\mathbf{x}; \Theta_{r}),
\end{equation}
where $\mathbf{x} \in \mathcal{X}$ and $\mathbf{z} \in \mathcal{Z}$. 
We further specify $\eta\big((\cdot,\cdot); \Theta_{s}\big)$ as an \textit{anomaly score learner} that uses a fully connected layer to learn linear pairwise relation features and the anomaly scores, taking the concatenation of the intermediate representations of each pair -- $\mathbf{z}_i$ and $\mathbf{z}_j$ -- as input:

\begin{equation}\label{eq:linear_mapping}
    \eta\big((\mathbf{z}_i,\mathbf{z}_j);\Theta_{s}\big) = \sum_{k=1}^{M}w^{o}_{k} z_{ik} + \sum_{l=1}^{M}w^{o}_{M+l} z_{jl} + w^{o}_{2M+1},
\end{equation}
where $\mathbf{z} \in \mathcal{Z}$ and $\Theta_{s} = \{\mathbf{w}^{o}\} $ in which $\{w^{o}_{1}, w^{o}_{2},\cdots,w^{o}_{2M}\}$ are weight parameters and $w^{o}_{2M+1}$ is a bias term. As shown in Fig. \ref{fig:framework}, 
PReNet uses a two-stream network with the shared weight parameters $\Theta_{r}$ to learn the representations $\mathbf{z}_i$ and $\mathbf{z}_j$. 
Thus, $\phi\big((\cdot,\cdot); \Theta\big)$ can be formally represented as 
\begin{equation}
    \phi\big((\mathbf{x}_i,\mathbf{x}_j);\Theta\big) = \eta\Big(\big(\psi(\mathbf{x}_i;\Theta_{r}),\psi(\mathbf{x}_j;\Theta_{r})\big);\Theta_s\Big),
\end{equation}
which can be trained in an end-to-end fashion. Note that the pairwise relation learned in Eqn. (\ref{eq:linear_mapping}) is a simple linear relation; learning more complex relations can be done by adding more layers with non-linear activation on top of the concatenated features, but it does not show clear advantages in our setting (see Table \ref{tab:ablation}).

PReNet then uses the pairwise ordinal class labels to optimize the pairwise relation-based anomaly scores. Particularly, it minimizes the difference between the prediction scores and the ordinal labels $y_{\mathbf{\{a,a\}}}$, $y_{\mathbf{\{a,u\}}}$ and $y_{\mathbf{\{u,u\}}}$. It is equivalent to learning to assign larger prediction scores to the anomaly-related (\ie, anomaly-anomaly and anomaly-unlabeled) instance pairs than the unlabeled-unlabeled pairs. 
Our loss is defined as below to guide the optimization:
\begin{equation}\label{eqn:loss}
    L\Big(\phi\big((\mathbf{x}_{i},\mathbf{x}_{j});\Theta\big), \yij \Big)= \bigg|\yij - \phi\big((\mathbf{x}_{i},\mathbf{x}_{j});\Theta\big)\bigg|.
\end{equation} 
The three-class labels $y_{\mathbf{\{a,a\}}}=8$, $y_{\mathbf{\{a,u\}}}=4$ and $y_{\mathbf{\{u,u\}}}=0$ are used by default to enforce a large margin among the anomaly scores of the three types of instance pairs. PReNet also works well with other value assignments as long as there are reasonably large margins among the ordinal labels (see Sec. \ref{subsec:furtheranalysis}). 
Lastly, PReNet is trained via:
\begin{multline}\label{eqn:obj}
    \argmin_{\Theta} \; \frac{1}{|\mathcal{B}|}\sum_{(\mathbf{x}_{i},\mathbf{x}_{j}, \yij) \in \mathcal{B}} \bigg|\yij - \phi\big((\mathbf{x}_{i},\mathbf{x}_{j});\Theta\big)\bigg|\\ + \lambda R(\Theta),
\end{multline}
where $\mathcal{B}$ is a sample batch from $\mathcal{P}$ and $R(\Theta)$ is a regularization term with hyperparameter $\lambda$. For each batch, $\frac{|\mathcal{B}|}{2}$ instance pairs are sampled from the $C_{\mathbf{\{u,u\}}}$ class and $\frac{|\mathcal{B}|}{4}$ instance pairs are respectively sampled from the $C_{\mathbf{\{a,a\}}}$ and $C_{\mathbf{\{a,u\}}}$ classes. This is equivalent to oversampling the two anomaly-related classes, $C_{\mathbf{\{a,a\}}}$ and $C_{\mathbf{\{a,u\}}}$, to avoid bias toward the $C_{\mathbf{\{u,u\}}}$ class due to the class imbalance.

\subsection{Anomaly Detection Using PReNet}

\noindent\textbf{Training.} Algorithm \ref{alg:prenet} presents the procedure of training PReNet. Step 1 first extends the data $\mathcal{X}$ into a set of instance pairs with ordinal class labels, $\mathcal{P}$. After a uniform \textit{Glorot} weight initialization in Step 2, PReNet performs stochastic gradient descent (SGD) based optimization to learn $\Theta$ in Steps 3-9 and obtains the optimized $\phi$ in Step 10. Particularly, stratified random sampling is used in Step 5 to ensure the sample balance of the three classes in $\mathcal{B}$, as discussed in Sec. \ref{subsubsec:scorelearning}.
Step 6 performs the forward propagation of the network and computes the loss. Step 7 then uses the loss to perform gradient descent steps.

\renewcommand{\algorithmicrequire}{\textbf{Input:}}
\renewcommand{\algorithmicensure}{\textbf{Output:}}
\begin{algorithm}
\small 
\caption{\textit{Training PReNet}}
\begin{algorithmic}[1]
\label{alg:prenet}
\REQUIRE $\mathcal{X} \in \mathbb{R}^{D}$ with $\mathcal{X}=\mathcal{U} \cup \mathcal{A}$ and $\emptyset=\mathcal{U} \cap \mathcal{A}$ 
\ENSURE $\phi: (\mathcal{X}, \mathcal{X}) \rightarrow \mathbb{R}$ - an anomaly score mapping
\STATE $\mathcal{P} \leftarrow$ Augment the training data with $\mathcal{U}$ and $\mathcal{A}$
\STATE Randomly initialize $\Theta$
\FOR{ $i = 1$ to $\mathit{n\_epochs}$}
    \FOR{ $j = 1$ to $\mathit{n\_batches}$}
        \STATE $\mathcal{B} \leftarrow$ Randomly sample $\mathit{b}$ data instance pairs from $\mathcal{P}$ \label{prenet:s5}
        \STATE $\mathit{loss} \leftarrow \frac{1}{b}\sum_{(\mathbf{x}_{i},\mathbf{x}_{j}, \yij) \in \mathcal{B}} \bigg|\yij\ - \phi\big((\mathbf{x}_{i},\mathbf{x}_{j});\Theta\big)\bigg| + \lambda R(\Theta)$    \label{prenet:s6}
        \STATE Perform a gradient descent step w.r.t. the parameters in $\Theta$
    \ENDFOR
\ENDFOR
\RETURN $\phi$
\end{algorithmic}
\end{algorithm}

\noindent \textbf{Inference.} During inference, given a test instance $\mathbf{x}_{k}$, PReNet first pairs it with data instances randomly sampled from $\mathcal{A}$ and $\mathcal{U}$, and then defines its anomaly score as
\begin{equation}\label{eqn:score2}
    \mathit{s}_{\mathbf{x}_{k}} = \frac{1}{2E} \left[ \sum_{i=1}^{E} \phi\big(\left(\mathbf{a}_{i},\mathbf{x}_{k}\right);\Theta^{*}\big) + \sum_{j=1}^{E} \phi\big((\mathbf{x}_{k},\mathbf{u}_{j});\Theta^{*}\big)\right],
\end{equation}
where $\Theta^{*}$ are the parameters of a trained $\phi$, and $\mathbf{a}_{i}$ and $\mathbf{u}_{j}$ are randomly sampled from the respective $\mathcal{A}$ and $\mathcal{U}$. $\mathit{s}_{\mathbf{x}_{k}}$ can be interpreted as an ensemble of the anomaly scores of a set of $\mathbf{x}_{k}$-oriented pairs. Due to the loss in Eqn. (\ref{eqn:loss}), $\mathit{s}_{\mathbf{x}_{k}}$ is optimized to be greater than $\mathit{s}_{\mathbf{x}_{k}^{\prime}}$ given $\mathbf{x}_{k}$ is an anomaly and $\mathbf{x}_{k}^{\prime}$ is a normal instance. 
PReNet can perform stably with a sufficiently large $E$ due to the law of large numbers ($E=30$ is used by default; see Sec. \ref{subsec:furtheranalysis} for other results).

\section{Theoretical Analysis}

\subsection{Pairwise Relation Feature Learning}\label{sec:transformativeaugmentation}
The random instance pairing module seamlessly leverages the two instance sets $\mathcal{A}$ and $\mathcal{U}$ to create large-scale proxy class labels to support pairwise relation feature learning. That is,
the sample size of the training data theoretically increases from $N+K$ in the original data space to $(N+K)^2$ for the pairwise relation learning,
including $K^2$ of $C_{\mathbf{\{a,a\}}}$ pairs, $2K\times N$ of $C_{\mathbf{\{a,u\}}}$ pairs and ${N^2}$ of $C_{\mathbf{\{u,u\}}}$ pairs (note that as set notion is used, $\mathbf{\{a,u\}}= \mathbf{\{u,a\}}$).
Such a large size helps build up the generalizability and then the performance of our detector. Note that PReNet uses a shared-weight two-stream network in $\psi(\cdot; \Theta_r)$, so the feature learning is still optimized on the $\mathcal{X}$ data space rather than the higher-order pairwise $\mathcal{P}$ space. This trick well supports the scale-up of the training sample size while adding no extra model complexity.
Further, the relation learning in Eqn. (\ref{eqn:obj}) enforces PReNet to discriminate the representations of anomaly-anomaly, anomaly-normal and normal-normal pairwise interactions (as $\mathcal{U}$ contains mostly normal data). This results in a joint learning of diverse patterns of abnormality, normality, and their interactions, avoiding the exclusive fitting of the seen abnormalities that may consequently overfit the seen abnormalities and fail to generalize to unseen abnormalities. 

\subsection{Robust Anomaly Score Learning}\label{sec:optimizedas}

This section analyzes the robustness of PReNet to $\epsilon$-contamination in the unlabeled data $\mathcal{U}$, where $\epsilon$ is the proportion of true anomalies in $\mathcal{U}$. Per the definition of anomaly, $\epsilon$ is typically small, \eg, $<2\%$. From the three-way modeling of anomaly-anomaly, anomaly-normal and normal-normal interactions, we can obtain the expectation of the pairwise relation proportions in each batch $\mathcal{B}$ based on uniformly random sampling. Let $\mathtt{a_T}$ and $\mathtt{n_T}$ indicate true anomaly and true
normal instances, respectively, we can then have the probability expectation
of each type of the interactions 
in $\mathcal{P}$ in Table \ref{tab:expect}.
\vspace{-0.2cm}
\begin{table}[htbp]
  \centering
\caption{Probability expectation of pairwise interactions}
\vspace{-0.2cm}
\begin{tabular}{|c|c|c|c|}
\cline{2-4} \cline{3-4} \cline{4-4}
\multicolumn{1}{c|}{} & {\scriptsize{}$\mathbf{\{\mathtt{a_{T}},\mathtt{a_{T}}\}}$} & {\scriptsize{}$\mathbf{\{\mathtt{a_{T}},\mathtt{n_{T}}\}}$} & {\scriptsize{}$\mathbf{\{\mathtt{n_{T}},\mathtt{n_{T}}\}}$}\tabularnewline
\hline
{\scriptsize{}$\mathbf{\{a,a\}}$} & {\scriptsize{}100\%} &  & \tabularnewline
\hline
{\scriptsize{}$\mathbf{\{a,u\}}$} & {\scriptsize{}$1*\epsilon$} & {\scriptsize{}$1-\epsilon$} & \tabularnewline
\hline
{\scriptsize{}$\mathbf{\{u,u\}}$} & {\scriptsize{}$\epsilon*\epsilon$} & {\scriptsize{}$2\epsilon\left(1-\epsilon\right)$} & {\scriptsize{}$\left(1-\epsilon\right)\left(1-\epsilon\right)$}\tabularnewline
\hline
\end{tabular}\label{tab:expect}
\end{table}

Considering the $\frac{1}{4}$,$\frac{1}{4}$, and $\frac{1}{2}$
sampling probability of anomaly and unlabeled pairs in $\mathcal{B}$,  there are $\frac{1}{4}+\frac{1}{4}\epsilon+\frac{1}{2}\epsilon^2$ from true anomaly-anomaly pairwise relations, and $\frac{1}{4}+\frac{3}{4}\epsilon-\epsilon^2$ from true anomaly-normal pairwise relations, and $\frac{1}{2}-\epsilon+\frac{1}{2}\epsilon^2$ from normal-normal pairwise relations. Their true expectation values when all the unlabeled cases are not true anomalies, are $\frac{1}{4}$, $\frac{1}{4}$, and $\frac{1}{2}$ respectively. Thus, a small percentage of the pairwise relations, $\left|\frac{1}{4}\epsilon+\frac{1}{2}\epsilon^{2}\right|+\left|\frac{3}{4}\epsilon-\epsilon^{2}\right|+\left|-\epsilon+\frac{1}{2}\epsilon^{2}\right|$=$2\epsilon - \epsilon^2$, would be expected to be noisy pairs. Considering the tolerance of the regression performance to about $5\%$ outliers~\cite{she2011outlier,yu2017robust}, PReNet can perform well when $\epsilon$ is reasonably small, $\leq 2.5\%$, which can often be satisfied for real-world anomaly detection problems. 
On the other hand, from the regression analysis, we can derive the following theorem for the expectation for different types of pairwise relation and their anomaly scores.

\begin{theorem}[Robustness to Anomaly Contamination]
Let $\epsilon\geq0$ be the anomaly contamination rate in $\mathcal{U}$, $y_{\mathbf{\{a,a\}}}=c_1$, $y_{\mathbf{\{a,u\}}}=c_2$, and $y_{\mathbf{\{u,u\}}}=c_3$ with $c_1>c_2>c_3\geq 0$, then for a given test instance $\mathbf{x}_k$, we have $\mathbb{E}\left[s_{\mathbf{x}_k} |\mathbf{x}_k\ is\ an\ anomaly \right] =\frac{c_1+c_2}{2}$, which is guaranteed to be greater than $\mathbb{E}\left[s_{\mathbf{x}_k} |\mathbf{x}_k\ is\ normal \right] =\frac{c_2+c_3+\epsilon (c_1+c_2)}{2}$ for $\epsilon <\frac{c_1-c_3}{c_1+c_2}$ (see Appendix \ref{subsec:proof} for the proof).
\end{theorem}

This theorem indicates that in PReNet a true anomaly is expected to have a larger anomaly score than normal instances when the contamination rate in $\mathcal{U}$ is not too large. That is, we have $\epsilon < \frac{2}{3}$ for our default setting: $c_1=8$, $c_2=4$ and $c_3=0$ (see Sec. \ref{subsubsec:scorelearning}), which is normally satisfied in real-world anomaly detection applications.

\section{Experiments}

\subsection{Datasets}

Multidimensional (or tabular) data is ubiquitous in real-world applications, so we focus on this type of publicly available datasets\footnote{See Appendix \ref{subsec:data} for more details about the used datasets.}.
To explicitly evaluate the performance of detecting seen/unseen anomalies, we have two groups of datasets from the literature
\cite{moustafa2015nb15,liu2012iforest,pang2019devnet}, including 12 datasets used for the detection of seen anomalies and another 28 datasets used for detecting unseen anomalies:

\vspace{0.1cm}
\noindent \textbf{Seen Anomaly Detection Datasets} As shown in Table \ref{tab:rocpr2}, 12 real-world datasets are used for the detection of seen anomalies, which are from diverse domains, \eg, cyber-attack detection, fraud detection, and disease risk assessment. Each dataset contains 0.2\%-15.0\% anomalies of the same class. To replicate the real-world scenarios where we have a few labeled seen anomalies and large unlabeled data, we first have a stratified split of each dataset into two subsets, with 80\% data as training data and the other 20\% data as a holdup test set. Since the unlabeled data is often anomaly-contaminated, we then combine some randomly selected anomalies with the normal training instances to form the unlabeled data $\mathcal{U}$. We further randomly sample a limited number of anomalies from the anomaly class to form the labeled anomaly set $\mathcal{A}$.

\vspace{0.1cm}
\noindent \textbf{Unseen Anomaly Detection Datasets} The 28 datasets for detecting unseen anomalies are presented in Table \ref{tab:unseen}. These datasets are derived from four intrusion attack datasets \textit{dos}, \textit{rec}, \textit{fuz} and \textit{bac} in Table \ref{tab:rocpr2}\footnote{The other eight datasets cannot be used in evaluating unseen anomaly detection as they are from different data sources and contain only one anomaly class.}, whose data instances are from the same data source and spanned by the same feature space. To guarantee that the anomalies in the test data are unseen during training, the anomaly class in one of these four datasets is held up for evaluation, while the anomalies in any combinations of the remaining three datasets are combined to form the pool of seen anomalies. 
We have 28 possible permutations under this setting, resulting in 28 datasets with different seen and/or unseen anomaly classes, as shown in Table \ref{tab:unseen}. During training, $\mathcal{A}$ contains the anomalies sampled from the pool of seen anomalies, while the test data is composed of the held-up unseen anomaly classes and the normal instances in the test set.

\subsection{Competing Methods and Their Settings}\label{sec:competing}
PReNet is compared with six state-of-the-art methods from several related areas,
including semi-supervised anomaly detectors, DevNet \cite{pang2019devnet,pang2021explainable} and one-class classifier Deep SAD (DSAD) \cite{ruff2019deep}, highly class-imbalanced (few-shot) classifier FSNet \cite{snell2017protonet} and its cost-sensitive variant cFSNet, and unsupervised anomaly detection methods iForest \cite{liu2012iforest} and REPEN \cite{pang2018repen} (REPEN represents unsupervised detectors that have a component to easily utilize any available anomaly data to train their models). 
Similar to \cite{pang2018repen,pang2019devnet}, we found empirically that all deep methods using a multilayer perceptron network architecture with one hidden layer perform better and more stably than using two or more hidden layers. 
Thus, following DevNet, one hidden layer with 20 neural units is used in all deep methods. The ReLu activation function $g(a) = \mathit{max}(0, a)$ is used. An $\ell_2$-norm regularizer with the hyperparameter setting $\lambda=0.01$ is applied to avoid overfitting. The RMSprop optimizer with the learning rate $0.001$ is used. All deep detectors are trained using 50 epochs, with 20 batches per epoch. 
Similar to PReNet, oversampling is also applied to the labeled anomaly set $\mathcal{A}$ to well train the deep detection models of DevNet, REPEN, DSAD, FSNet and cFSNet. iForest with recommended settings \cite{liu2012iforest} is used as a baseline here. 

We also compare PReNet with XGBOD \cite{zhao2018xgbod}, PUMAD \cite{ju2020pumad}, and FEAWAD \cite{zhou2022feature}. The comparison results are given in Appendix \ref{sec:additional_results} due to space limitation.

\subsection{Performance Evaluation Metrics}

Two popular metrics -- the Area Under Receiver Operating Characteristic Curve (AUC-ROC) and Area Under Precision-Recall Curve (AUC-PR) -- are used.
A larger AUC-ROC/AUC-PR reflects better performance. AUC-ROC that summarizes the curve of true positives against false positives often presents an overoptimistic view of the performance, whereas AUC-PR is more practical as it summarizes the precision and recall w.r.t. the anomaly class exclusively \cite{boyd2013aucpr}. 
The reported results are averaged values over 10 independent runs. The paired \textit{Wilcoxon} signed-rank \cite{woolson2007wilcoxon} is used to examine the statistical significance of PReNet against its competing methods.

\subsection{Detection of Seen Anomalies}\label{subsec:knownanomalies}

\textbf{Effectiveness of PReNet using small anomaly examples.} We first evaluate PReNet on detecting seen anomalies in 12 real-world datasets. A consistent anomaly contamination rate and the same number of labeled anomalies are used across all datasets to gain insights into the performance in different real-life applications. Since anomalies are typically rare instances, the number of labeled anomalies available per data is set to 60, \ie, $|\mathcal{A}|=60$, and the anomaly contamination rate is set to 2\% by default.

\begin{table*}[htbp]
  \centering
\caption{Seen anomaly detection results. `Size' is the data size. $D$ is the dimension. `1M' denotes \textit{news20} has 1,355,191 features.}
\scalebox{0.81}{
    \begin{tabular}{lccccccccc|ccccccc}\hline\hline
\multicolumn{3}{c}{\textbf{Data Statistics}} & \multicolumn{7}{c|}{\textbf{AUC-PR Results}}& \multicolumn{7}{c}{\textbf{AUC-ROC Results}}\\\hline
\textbf{Data} & \centering \textbf{Size} & \centering$D$ & \centering\textbf{PReNet} & \centering\textbf{DevNet} & \centering\textbf{DSAD} & \centering\textbf{FSNet} & \centering \textbf{cFSNet} & \centering \textbf{REPEN} & \textbf{iForest}& \centering\textbf{PReNet} & \centering\textbf{DevNet} & \centering\textbf{DSAD} & \centering\textbf{FSNet} & \centering \textbf{cFSNet} & \centering \textbf{REPEN} & \textbf{iForest}\\\hline
    donors & 619,326 & 10    & \textbf{1.000} & 0.997 & 0.806 & 0.995 & 0.998 & 0.520  & 0.222 & \textbf{1.000} & \textbf{1.000} & 0.993 & 0.999 & 1.000     & 0.976 & 0.875 \\
    census & 299,285 & 500   & \textbf{0.356} & 0.345 & 0.330  & 0.197 & 0.197 & 0.173 & 0.078 & \textbf{0.862} & 0.861 & 0.858 & 0.759 & 0.759 & 0.822 & 0.634 \\
    fraud & 284,807 & 29    & 0.689 & 0.693 & \textbf{0.695} & 0.157 & 0.169 & 0.678 & 0.261 & 0.980  & \textbf{0.981} & 0.980  & 0.776 & 0.762 & 0.974 & 0.946 \\
    celeba & 202,599 & 39    & \textbf{0.309} & 0.306 & 0.306 & 0.103 & 0.112 & 0.166 & 0.060  & 0.960  & \textbf{0.961} & 0.958 & 0.855 & 0.855 & 0.899 & 0.686 \\
    dos   & 109,353 & 196   & 0.900   & 0.900   & \textbf{0.910} & 0.826 & 0.836 & 0.461 & 0.266 & 0.949 & 0.948 & \textbf{0.956} & 0.927 & 0.935 & 0.890  & 0.762 \\
    rec   & 106,987 & 196   & 0.767 & 0.760  & \textbf{0.772} & 0.650  & 0.679 & 0.302 & 0.132 & 0.966 & 0.962 & \textbf{0.971} & 0.926 & 0.936 & 0.829 & 0.534 \\
    fuz   & 96,000 & 196   & \textbf{0.170} & 0.136 & 0.133 & 0.146 & 0.146 & 0.075 & 0.039 & \textbf{0.882} & 0.873 & 0.875 & 0.858 & 0.858 & 0.789 & 0.548 \\
    bac   & 95,329 & 196   & \textbf{0.890} & 0.863 & 0.862 & 0.618 & 0.618 & 0.117 & 0.050  & \textbf{0.976} & 0.968 & 0.945 & 0.950  & 0.950  & 0.882 & 0.741 \\
    w7a   & 49,749 & 300   & \textbf{0.496} & 0.408 & 0.117 & 0.098 & 0.098 & 0.112 & 0.023 & \textbf{0.883} & 0.882 & 0.802 & 0.767 & 0.767 & 0.733 & 0.413 \\
    campaign & 41,188 & 62    & \textbf{0.470} & 0.426 & 0.386 & 0.255 & 0.255 & 0.333 & 0.313 & \textbf{0.880} & 0.858 & 0.803 & 0.684 & 0.684 & 0.740  & 0.723 \\
    news20 & 10,523 & 1M    & \textbf{0.652} & 0.632 & 0.329 & 0.105 & 0.116 & 0.222 & 0.035 & 0.956 & \textbf{0.960} & 0.909 & 0.686 & 0.690  & 0.869 & 0.333 \\
    thyroid & 7,200 & 21    & \textbf{0.298} & 0.280  & 0.205 & 0.149 & 0.175 & 0.108 & 0.144 & \textbf{0.781} & 0.767 & 0.713 & 0.590  & 0.594 & 0.613 & 0.679 \\
    \multicolumn{3}{r}{\textbf{Average}}& \centering \textbf{0.583} & \centering 0.562 & \centering 0.488 & \centering 0.358 &  \centering 0.367 &  \centering 0.272 & 0.135 & \centering \textbf{0.923} & \centering 0.918 & \centering 0.897 & \centering 0.815 & \centering 0.816  & \centering 0.835 & 0.656\\\hline
\multicolumn{3}{r}{\textbf{P-value}}& \centering - & \centering 0.005 & \centering 0.016 & \centering 0.001 &  \centering 0.001 &\centering 0.001 & 0.001& \centering - & \centering 0.075 & \centering 0.025 & \centering 0.001 & \centering 0.001 & \centering 0.001 & 0.001 \\\hline\hline
    \end{tabular}%
  \label{tab:rocpr2}%
  }
\end{table*}%

\begin{figure*}[t!]
  \centering
    \includegraphics[width=0.8\textwidth]{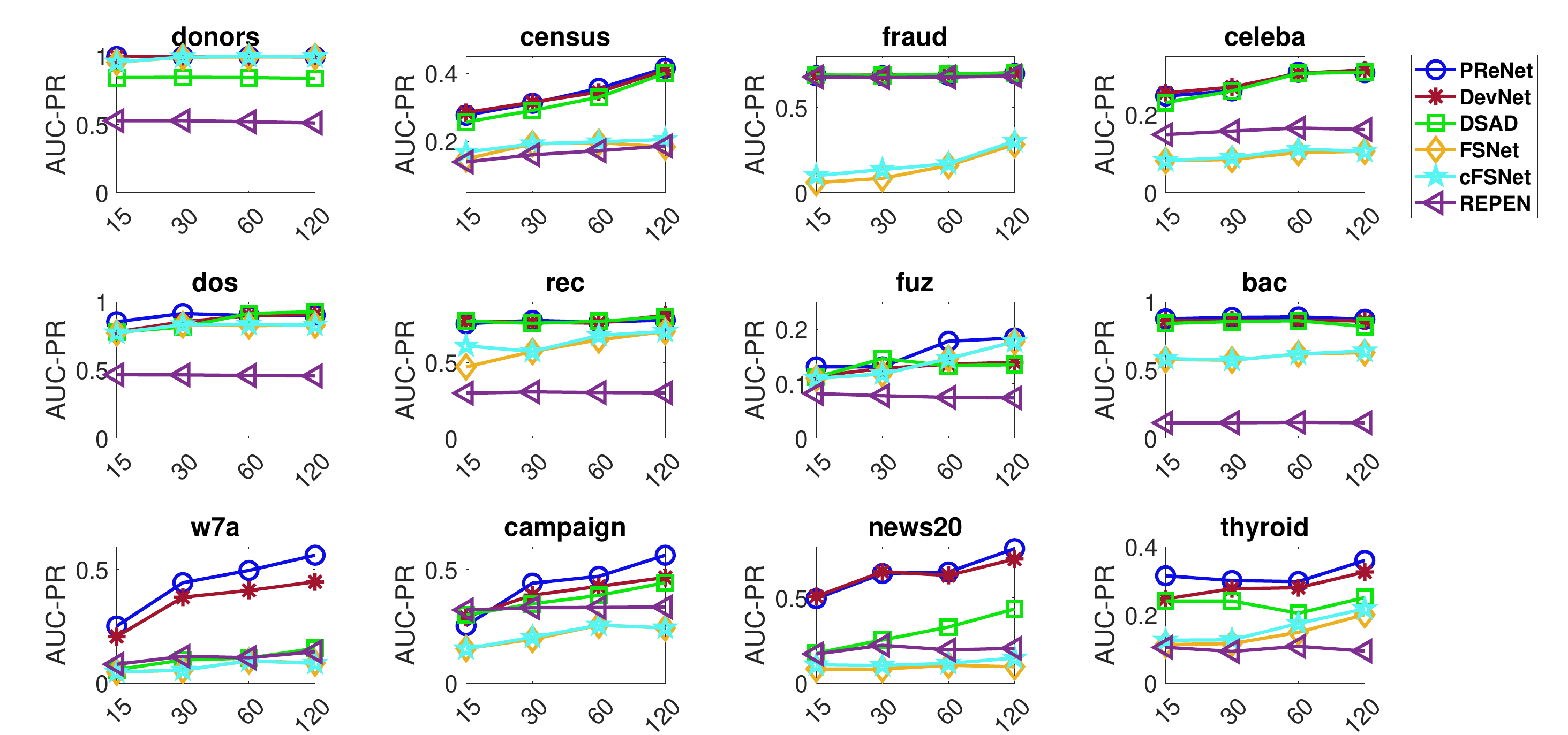}
  \caption{AUC-PR results of seen anomaly detection w.r.t. the number of labeled anomalies}
  \label{fig:knownouliers}
\end{figure*}

The results on the 12 datasets are shown in Table \ref{tab:rocpr2}. In AUC-PR, PReNet performs substantially better than, or comparably well to, all competing methods across the 12 datasets. On average, PReNet improves all five competing methods by a large margin, \ie, DevNet (3.7\%), DSAD (19.6\%), FSNet (54.0\%), cFSNet (59.1\%), REPEN (114.2\%), and iForest (331.85\%), which are all statistically significant at the 95\%/99\% confidence level. Particularly, compared to the top two contenders, PReNet significantly outperforms DevNet on six datasets, with improvement ranging from 3\%-6\% on \textit{census}, \textit{bac}, \textit{news20} and \textit{thyroid}, and up to 10\%-20\% on \textit{campaign} and \textit{w7a}; they perform comparably well on the rest of six datasets; PReNet performs significantly better than DSAD on eight datasets, achieving 20\%-320\% improvement on six datasets, including \textit{donors}, \textit{fuz}, \textit{w7a}, \textit{campaign}, \textit{news20} and \textit{thyroid}, and they perform comparably well on the rest of four datasets. 
In terms of AUC-ROC, PReNet outperforms DevNet at the 90\% confidence level, and performs significantly better than DSAD (2.9\%), FSNet (11.7\%), cFSNet (13.1\%), REPEN (10.6\%) and iForest (40.7\%) at the 95\%/99\%  confidence level.

\begin{table*}[htbp]
  \centering
   \caption{Unseen anomaly detection results. The models are trained with `seen' anomaly classes to detect `unseen' anomalies.}
\scalebox{0.82}{
    \begin{tabular}{lcccccccc|ccccccc}
    \hline\hline
     \multicolumn{2}{c}{\textbf{Anomaly Class}}      & \multicolumn{7}{c|}{\textbf{AUC-PR Results}} & \multicolumn{7}{c}{\textbf{AUC-ROC Results}} \\\hline
    \textbf{Seen }   & \textbf{Unseen}  & \textbf{PReNet} &  \textbf{DevNet}  & \textbf{DSAD}    & \textbf{FSNet}  & \textbf{cFSNet} & \textbf{REPEN}& \textbf{iForest} & \textbf{PReNet} &  \textbf{DevNet}  & \textbf{DSAD}    & \textbf{FSNet}  & \textbf{cFSNet} & \textbf{REPEN}& \textbf{iForest}    \\\hline
        dos    &  bac  &  \textbf{0.908} &  \textbf{0.908} & 0.890 & 0.772 & 0.841 & 0.590  & 0.050  &  \textbf{0.958} & 0.956 & 0.924 & 0.932 & 0.943 & 0.889 & 0.741 \\
        dos, fuz  &  bac  &  \textbf{0.889} & 0.886 & 0.879 & 0.555 & 0.679 & 0.438 & 0.050  &  \textbf{0.969} & 0.967 & 0.918 & 0.921 & 0.941 & 0.886 & 0.741 \\
        fuz  &  bac  & 0.503 & 0.219 & 0.266 & 0.366 &  \textbf{0.524} & 0.500   & 0.050  & 0.869 & 0.794 & 0.812 & 0.872 & 0.896 &  \textbf{0.899} & 0.741 \\
        rec  &  bac  &  \textbf{0.752} & 0.541 & 0.583 & 0.550  & 0.637 & 0.474 & 0.050  &  \textbf{0.965} & 0.900   & 0.902 & 0.885 & 0.921 & 0.846 & 0.741 \\
        rec, dos  &  bac  & 0.834 &  \textbf{0.845} & 0.833 & 0.48  & 0.537 & 0.548 & 0.050  & 0.977 &  \textbf{0.980} & 0.978 & 0.924 & 0.942 & 0.882 & 0.741 \\
        rec, dos, fuz  &  bac  &  \textbf{0.706} & 0.631 & 0.623 & 0.368 & 0.368 & 0.500   & 0.050  &  \textbf{0.971} & 0.953 & 0.952 & 0.916 & 0.916 & 0.886 & 0.741 \\
        rec, fuz  &  bac  &  \textbf{0.711} & 0.342 & 0.317 & 0.312 & 0.387 & 0.436 & 0.050  &  \textbf{0.969} & 0.891 & 0.89  & 0.865 & 0.902 & 0.874 & 0.741 \\
        bac  &  dos    & 0.938 & 0.943 &  \textbf{0.961} & 0.93  & 0.944 & 0.769 & 0.266 & 0.906 & 0.915 &  \textbf{0.945} & 0.926 & 0.933 & 0.881 & 0.762 \\
        bac, fuz  &  dos    &  \textbf{0.932} & 0.761 & 0.772 & 0.714 & 0.803 & 0.801 & 0.266 &  \textbf{0.958} & 0.889 & 0.887 & 0.872 & 0.911 & 0.924 & 0.762 \\
        fuz  &  dos    & 0.811 & 0.644 & 0.68  & 0.774 & 0.803 &  \textbf{0.846} & 0.266 & 0.855 & 0.792 & 0.805 & 0.826 & 0.846 &  \textbf{0.921} & 0.762 \\
        rec  &  dos    &  \textbf{0.928} & 0.846 & 0.855 & 0.798 & 0.831 & 0.771 & 0.266 &  \textbf{0.938} & 0.883 & 0.887 & 0.825 & 0.846 & 0.875 & 0.762 \\
        rec, bac  &  dos    &  \textbf{0.891} & 0.870  & 0.871 & 0.686 & 0.762 & 0.742 & 0.266 &  \textbf{0.944} & 0.932 & 0.931 & 0.872 & 0.887 & 0.904 & 0.762 \\
        rec, bac, fuz  &  dos    &  \textbf{0.835} & 0.610  & 0.699 & 0.572 & 0.627 & 0.641 & 0.266 &  \textbf{0.940} & 0.861 & 0.886 & 0.821 & 0.858 & 0.890  & 0.762 \\
        rec, fuz  &  dos    &  \textbf{0.883} & 0.718 & 0.673 & 0.670  & 0.764 & 0.749 & 0.266 &  \textbf{0.939} & 0.874 & 0.868 & 0.834 & 0.88  & 0.903 & 0.762 \\
        bac  &  fuz  & 0.418 &  \textbf{0.420} & 0.250  & 0.374 & 0.383 & 0.251 & 0.039 &  \textbf{0.752} & 0.743 & 0.364 & 0.697 & 0.734 & 0.695 & 0.548 \\
        dos    &  fuz  & 0.418 & 0.427 & 0.325 & 0.288 & 0.324 &  \textbf{0.476} & 0.039 & 0.708 & 0.737 & 0.508 & 0.606 & 0.708 &  \textbf{0.783} & 0.548 \\
        dos, bac  &  fuz  & 0.375 & 0.371 & 0.322 & 0.273 & 0.301 &  \textbf{0.463} & 0.039 & 0.842 & 0.833 & 0.646 & 0.837 &  \textbf{0.860} & 0.828 & 0.548 \\
        rec  &  fuz  &  \textbf{0.462} & 0.418 & 0.419 & 0.410  & 0.427 & 0.408 & 0.039 &  \textbf{0.878} & 0.872 & 0.872 & 0.843 & 0.838 & 0.777 & 0.548 \\
        rec, bac  &  fuz  & 0.315 & 0.311 & 0.314 & 0.260  & 0.255 &  \textbf{0.319} & 0.039 & 0.879 & 0.879 &  \textbf{0.880} & 0.838 & 0.824 & 0.797 & 0.548 \\
        rec, bac, dos  &  fuz  & 0.294 & 0.246 & 0.249 & 0.206 & 0.189 &  \textbf{0.349} & 0.039 &  \textbf{0.885} & 0.878 & 0.878 & 0.846 & 0.853 & 0.832 & 0.548 \\
        rec, dos  &  fuz  & 0.349 & 0.375 & 0.366 & 0.276 & 0.306 &  \textbf{0.434} & 0.039 & 0.850  &  \textbf{0.889} & 0.871 & 0.822 & 0.837 & 0.800   & 0.548 \\
        bac  &  rec  &  \textbf{0.892} & 0.890  & 0.576 & 0.689 & 0.718 & 0.592 & 0.132 &  \textbf{0.928} & 0.926 & 0.489 & 0.693 & 0.713 & 0.741 & 0.534 \\
        bac, fuz  &  rec  &  \textbf{0.876} & 0.804 & 0.831 & 0.672 & 0.692 & 0.554 & 0.132 &  \textbf{0.958} & 0.943 & 0.947 & 0.879 & 0.897 & 0.822 & 0.534 \\
        dos    &  rec  &  \textbf{0.849} & 0.846 & 0.739 & 0.615 & 0.718 & 0.686 & 0.132 &  \textbf{0.867} & 0.865 & 0.677 & 0.636 & 0.727 & 0.778 & 0.534 \\
        dos, bac  &  rec  &  \textbf{0.768} & 0.732 & 0.670  & 0.618 & 0.644 & 0.597 & 0.132 &  \textbf{0.908} & 0.891 & 0.724 & 0.859 & 0.868 & 0.805 & 0.534 \\
        dos, bac, fuz  &  rec  &  \textbf{0.719} & 0.716 & 0.629 & 0.540  & 0.586 & 0.447 & 0.132 &  \textbf{0.907} &  \textbf{0.907} & 0.820  & 0.871 & 0.892 & 0.783 & 0.534 \\
        dos, fuz  &  rec  &  \textbf{0.788} & 0.772 & 0.615 & 0.631 & 0.644 & 0.542 & 0.132 &  \textbf{0.899} & 0.885 & 0.663 & 0.842 & 0.863 & 0.800   & 0.534 \\
        fuz  &  rec  &  \textbf{0.797} & 0.718 & 0.727 & 0.760  & 0.785 & 0.672 & 0.132 &  \textbf{0.890} & 0.855 & 0.863 & 0.871 & 0.885 & 0.812 & 0.534 \\
     \multicolumn{2}{r}{\textbf{Average}}    & \textbf{0.709} & 0.636 & 0.605 & 0.542  & 0.589  & 0.557 &   - & \textbf{0.904} & 0.882  & 0.814 & 0.837  & 0.861  & 0.840 &  -\\\hline
\multicolumn{2}{r}{\textbf{P-value}}& \centering - & \centering 0.001 & \centering 0.000 & \centering 0.000 & \centering 0.000 &0.000 & -& \centering - & \centering 0.001 & \centering 0.000 & \centering 0.000 & \centering 0.000 & 0.000&  -\\\hline\hline
    \end{tabular}%
  \label{tab:unseen}%
  }
\end{table*}%

\vspace{0.1cm}
\noindent \textbf{Using less/more labeled anomaly examples.} We further examine PReNet by evaluating its performance w.r.t. different numbers of labeled anomalies, ranging from 15 to 120, with the contamination rate fixed to 2\%. The AUC-PR results are shown in Fig. \ref{fig:knownouliers}. iForest is omitted as it does not use labeled data. The results of all methods generally increases with labeled data size. However, The increased anomalies do not always help due to the heterogeneous anomalous behaviors taken by different anomalies. PReNet is more stable in the increasing trend. Consistent with the results in Table \ref{tab:rocpr2}, PReNet still significantly outperforms its state-of-the-art competing methods with varying numbers of anomaly examples. Particularly, PReNet demonstrates the most sample-efficient learning capability. Impressively, PReNet can be trained with 50\%-75\% less labeled anomalies but achieves much better, or comparably good, AUC-PR than the best contender DevNet on multiple datasets like \textit{dos}, \textit{fuz}, \textit{w7a} and \textit{campaign}; and it is trained with 87.5\% less labeled data while obtains substantially better performance than the second-best contender DSAD on \textit{donors}, \textit{w7a}, \textit{campaign}, \textit{news20} and \textit{thyroid}. Similar observations apply to FSNet, cFSNet and REPEN.

\subsection{Detection of Unseen Anomalies}

\textbf{Generalizing to \underline{unseen} anomaly classes using small examples of \underline{seen} anomaly classes.} This section evaluates the detectors that are trained with only seen anomaly classes to detect unseen anomaly classes on the 28 datasets. Similarly to Sec. \ref{subsec:knownanomalies}, the anomaly contamination rate of 2\% and $|\mathcal{A}|=60$ are used here. The results are presented in Table \ref{tab:unseen}, in which iForest that is insensitive to the change is used as baseline. The AUC-PR results show that PReNet outperforms all the five competing methods by substantial margins on the 28 datasets. On average, PReNet improves DevNet by more than 11\%, DSAD by 17\%, FSNet by 30\%, cFSNet by 20\% and REPEN by 27\%. It is impressive that, compared to the best competing method DevNet, PReNet achieves 20\%-130\% AUC-PR improvement on eight datasets, including 20\%-40\% improvement on `\textit{rec $\rightarrow$ bac}', `\textit{rec,fuz $\rightarrow$ dos}', `\textit{rec, bac, fuz $\rightarrow$ dos}', `\textit{bac, fuz $\rightarrow$ dos}', `\textit{fuz $\rightarrow$ dos}', `\textit{rec, bac, dos $\rightarrow$ fuz}', and over 100\% improvement on `\textit{rec, fuz $\rightarrow$ bac}' and `\textit{fuz $\rightarrow$ bac}'. The improvement over the other four contenders is more substantial than that over DevNet. 
All these improvements are statistically significant at the 99\% confidence level. PReNet gains similar superiority in AUC-ROC as well.

\begin{figure*}[h!]
  \centering
    \includegraphics[width=0.8\textwidth]{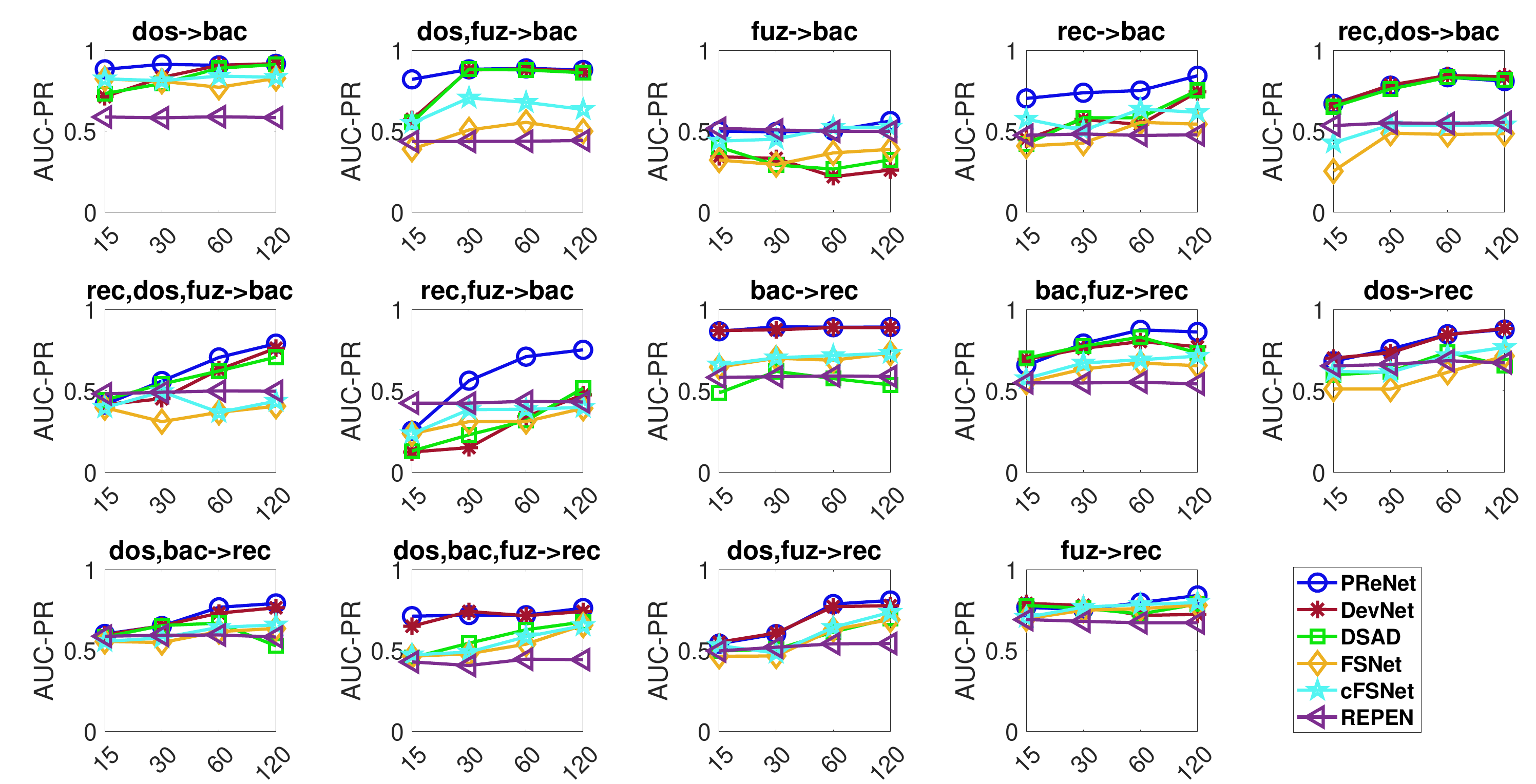}
  \caption{AUC-PR w.r.t. \# of labeled anomalies. `A -$>$ B' means the models trained with attacks `A' to detect unseen attacks `B'.}
  \label{fig:unknownouliers}
\end{figure*}

\vspace{0.1cm}
\noindent \textbf{Detection of unseen anomaly classes with less/more seen anomalies.} We examine this question on 14 unknown anomaly datasets where all methods work relatively well compared to the other 14 datasets. The AUC-PR results are presented in Fig. \ref{fig:unknownouliers}. It is interesting that most detectors gain improved performance in detecting unseen anomalies when more seen anomalies are given. This may be due to that more seen anomaly examples help better train the detection models, enabling a better anomaly discriminability. The superiority of PReNet here is consistent with that in Table \ref{tab:unseen}. PReNet remains the most sample-efficient method, and can perform substantially better than the best competing methods even when the PReNet model is trained with 50\%-87.5\% less labeled data.

\subsection{Further Analysis of PReNet}\label{subsec:furtheranalysis}
\begin{figure}[h!]
  \centering
    \includegraphics[width=0.49\textwidth]{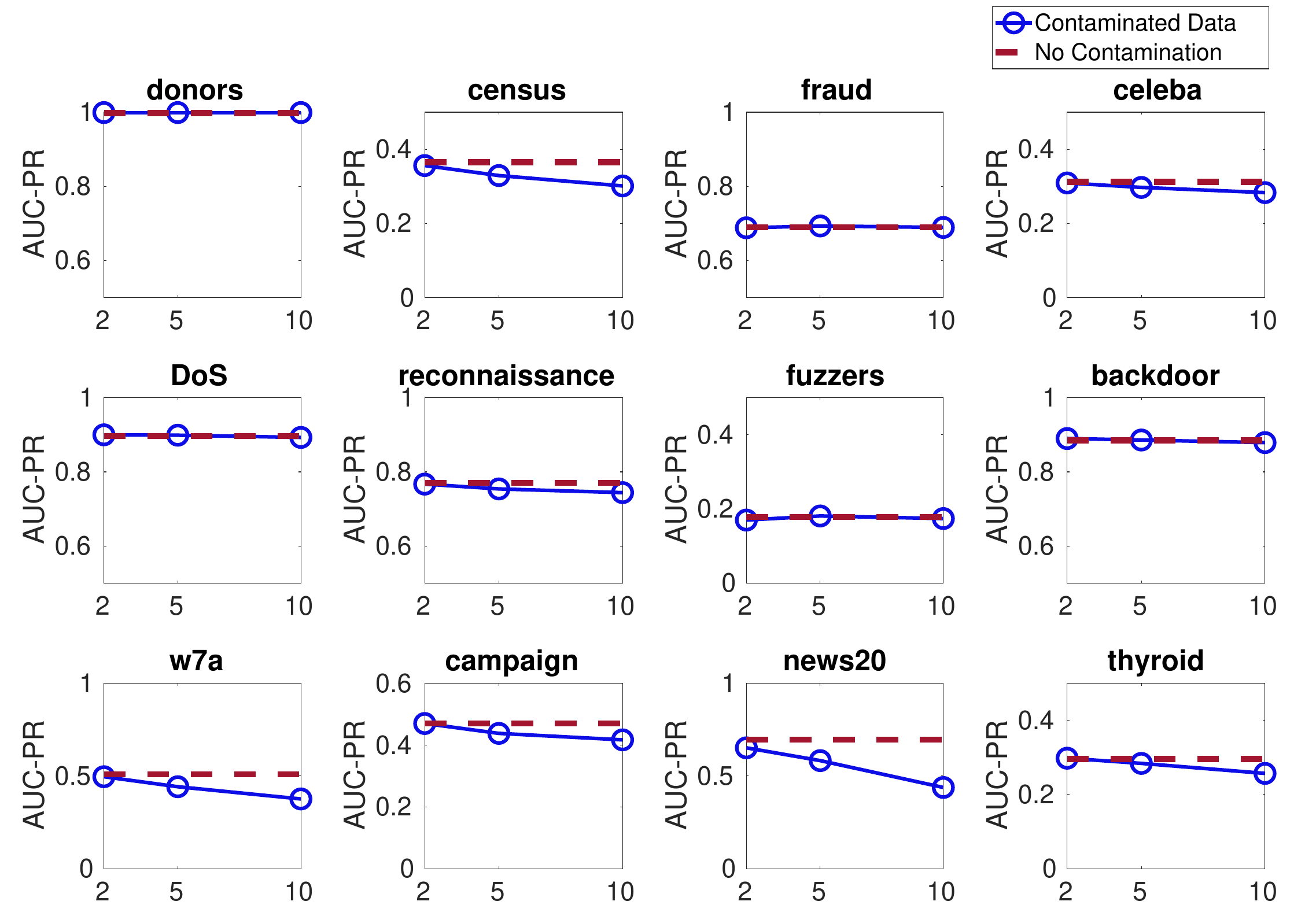}
  \caption{AUC-PR w.r.t. different contamination rates (\%). }
  \label{fig:contrate}
\end{figure}

\noindent \textbf{Robustness w.r.t. anomaly contamination.} We investigate this robustness by using different anomaly contamination rates, \{0\%, 2\%, 5\%, 10\%\}, with $|\mathcal{A}|=60$ fixed. 
The AUC-PR results for this experiment are presented in Fig. \ref{fig:contrate}. PReNet performs generally stably on all datasets with the contamination rate below 10\%, except \textit{news20} that contains over one millions features and may therefore require better relation learning designs to achieve good robustness w.r.t. a large contamination rate.

\vspace{0.1cm}
\noindent \textbf{Ablation study.} In Table \ref{tab:ablation}, PReNet is compared to its four ablated variants to evaluate the importance of each module:

\begin{itemize}
    \item \textbf{Three-way relation modeling.} To examine the importance of learning diverse pairwise patterns via three-way relation modeling, we compare PReNet with its Two-way Relation Modeling (TRM) variant that 
    learns two-way relations only, \ie, to discriminate $y_{\mathbf{\{a,a\}}|\mathbf{\{a,u\}}}$ from $y_{\mathbf{\{u,u\}}}$. PReNet outperforms than TRM on nearly all datasets, resulting in large average improvement (5.6\%). This indicates that learning more diverse patterns helps obtain better generalization.
    \item \textbf{Relation features.} Can PReNet perform similarly well by learning features of individual instances rather than pairwise relation features? The answer is negative. Compared to iPReNet in Table \ref{tab:ablation} that takes individual instances as input and learns to discriminate seen anomalies from unlabeled data, PReNet performs better by a large margin on 11 out of 12 datasets.
    \item \textbf{Feature learning layer $\psi$}. To evaluate the importance of intermediate feature representations, PReNet is compared with LDM that removes the hidden layers of PReNet (\ie, the $\psi$ function) and learns a Linear Direct Mapping (LDM) from the original data space to anomaly scores. The results show that eliminating the feature layer leads to over 10\% loss in the average AUC-PR performance.
    \item \textbf{Deeper neural network}. We also explore the possibility of learning a deeper network in PReNet. A2H is a variant of PReNet, which deepens PReNet with additional two hidden (A2H) layers. Each added layer is regularized by an $\ell_2$-norm regularizer and a dropout layer of 0.5 dropout rate. 
    Although A2H performs well on some datasets like \textit{rec} and \textit{thyroid}, it fails on most of the other datasets. As a result, PReNet can still gain large average improvement over A2H. Thus, the default architecture used in PReNet is generally recommended. 
    \item \textbf{Non-linear relation learning}. We compare PReNet with a variant of Non-linear Pairwise Relation (NPR) learning that adds a non-linear layer in-between the the concatenated features and the anomaly scoring layer. Similar to A2H, NRP can work better than PReNet on a few cases, but it is often too complex and has an overfitting problem on most datasets.  
\end{itemize}

\begin{table}[t!]
\caption{AUC-PR results of ablation study.}
\begin{center}
\scalebox{0.85}{
\begin{tabular}{lcccccc}
\hline
\textbf{Data} & \centering \textbf{PReNet} &  \centering \textbf{TRM} &  \centering \textbf{iPReNet} &  \centering \textbf{LDM} &  \textbf{A2H} & \textbf{NPR} \\ \hline
donors	&	\textbf{1.000}	&	0.995	&	0.999	&	0.974	&	\textbf{1.000} & \textbf{1.000}\\
census	&	\textbf{0.356}	&	0.324	&	0.325	&	0.355	&	0.283 & 0.310\\
fraud	&	0.689	&	0.692	&	0.701	&	0.662	&	\textbf{0.708} & 0.683\\
celeba	&	\textbf{0.309}	&	0.308	&	0.299	&	0.294	&	0.299 & 0.286\\
dos	&	0.900	&	0.887	&	0.887	&	0.839	&	0.877& \textbf{0.924}\\
rec	&	0.767	&	0.751	&	0.75	&	0.647	&	\textbf{0.877} & 0.814\\
fuz	&	0.170	&	0.147	&	0.151	&	0.163	&	\textbf{0.178} & 0.154 \\
bac	&	0.890	&	0.879	&	0.879	&	0.805	&	0.863& \textbf{0.898}\\
w7a	&	\textbf{0.496}	&	0.467	&	0.482	&	0.406	&	0.415 & 0.393\\
campaign	&	\textbf{0.470}	&	0.423	&	0.402	&	0.406	&	0.246&0.370\\
news20	&	\textbf{0.652}	&	0.481	&	0.625	&	0.552	&	0.618 & 0.607\\
thyroid	&	0.298	&	0.269	&	0.248	&	0.201	&	\textbf{0.411} & 0.341\\\hline
\textbf{P-value}& - &  0.002&  0.003 &  0.001 & 0.413 & 0.328\\\hline
\end{tabular}
}
\end{center}
\label{tab:ablation}
\end{table}

\vspace{0.1cm}
\noindent \textbf{Sensitivity test.} We evaluate the sensitivity of PReNet w.r.t. three key hyperparameters: $y_{\{\xvec_i, \xvec_j\}}$ and $\lambda$ in Eqn. (\ref{eqn:obj}), and $E$ in Eqn. (\ref{eqn:score2}).

\begin{figure}[h!]
  \centering
    \includegraphics[width=0.4\textwidth]{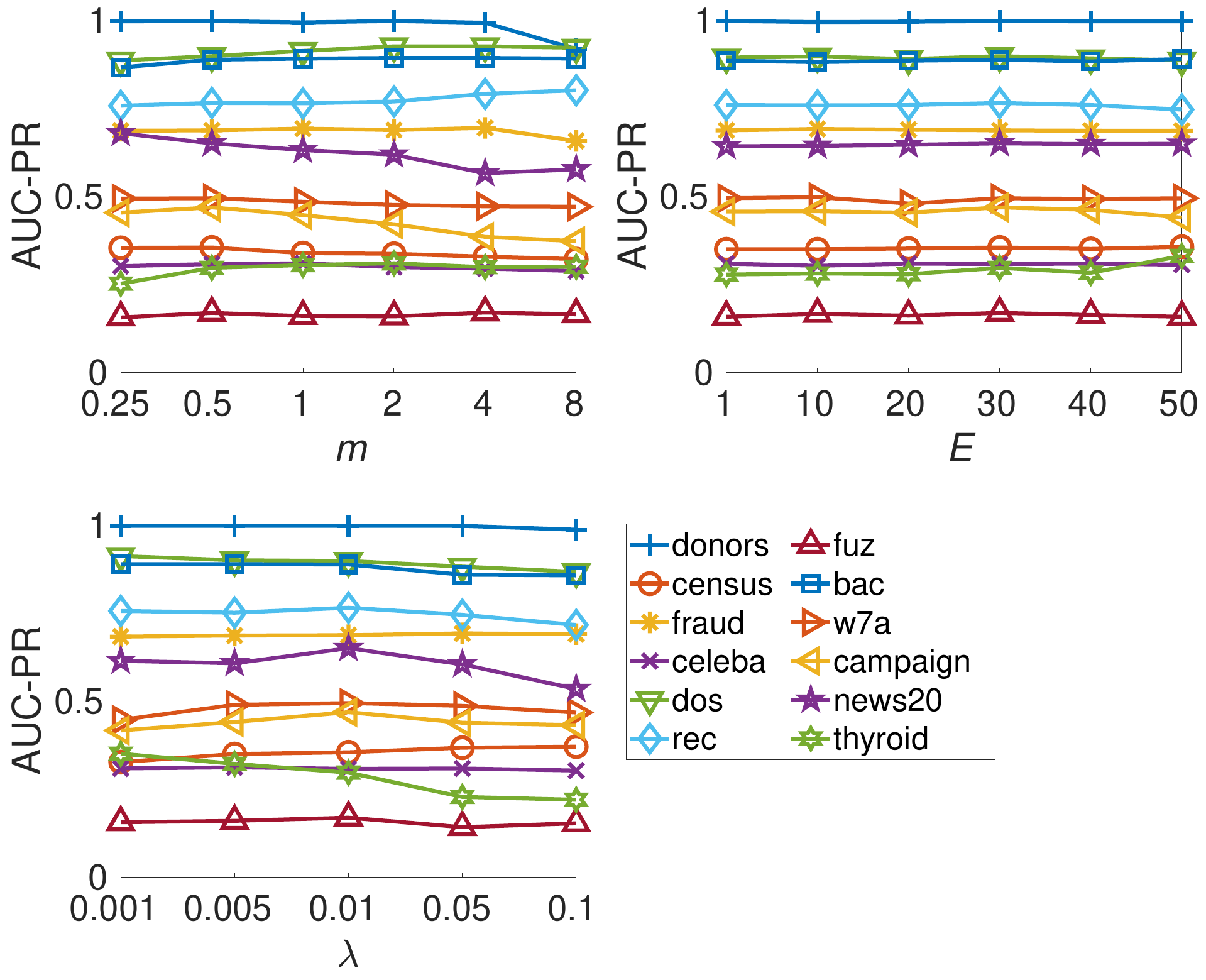}
  \caption{AUC-PR of PReNet w.r.t. three hyperparameters.}
  \label{fig:contrate2}
\end{figure}

\begin{itemize}
    \item \textbf{Sensitivity w.r.t. pairwise class labels $y_{\{\xvec_i, \xvec_j\}}$}. This section examines the sensitivity of PReNet w.r.t. the synthetic ordinal pairwise relation class labels. We fix the ordinal label for $y_{\{\uvec,\uvec\}}$ to be zero, \ie, $c_3=0$, and the same margin is set between $y_{\{\uvec,\uvec\}}$ and $y_{\{\avec,\uvec\}}$ pairs,  and between $y_{\{\avec,\uvec\}}$ and $y_{\{\avec,\avec\}}$ pairs, \ie, $(c_2-c_3)=(c_1-c_2)=m$. We test the sensitivity w.r.t. different values of the margin $m$. The AUC-PR results are shown in Figure \ref{fig:contrate2}. It is clear that PReNet is generally robust to different margin values. PReNet performs well even when setting a rather small margin, \eg, $m=0.25$. 
    Larger margins are generally more desired, especially in some challenging datasets such as \textit{thyroid} and \textit{dos}. 

    \item \textbf{Sensitivity w.r.t. ensemble size $E$}. This section investigates the sensitivity of PReNet w.r.t. the ensemble size $E$ in Eqn. (\ref{eqn:score2}). 
    The AUC-PR results are shown in Figure \ref{fig:contrate2}. PReNet performs very stably across all the 12 datasets. PReNet using small $E$ performs similarly well as that using a large $E$, indicating that highly discriminative features are learned in PReNet. 
    Increasing $E$ may offer better detection accuracy on some datasets,
    but the improvement is often marginal.

    \item \textbf{Sensitivity w.r.t. regularization parameter $\lambda$}. This part investigates the sensitivity of PReNet w.r.t. a wide range of $\lambda$ settings, $\lambda=\{0.001, 0.005, 0.01, 0.05, 0.1\}$, in Eqn. (5). 
    The AUC-PR results are shown in Figure \ref{fig:contrate2}. PReNet is generally robust w.r.t. different $\lambda$ values on all the 12 datasets, especially when $\lambda$ is chosen in the range $[0.001,0.01]$. When increasing $\lambda$ to larger values such as 0.05 or 0.1, the AUC-PR of PReNet decreases on a few datasets, \eg, \textit{rec}, \textit{bac}, \textit{news20} and \textit{thyroid}. This may be due to that given the limited number of labeled anomaly data, enforcing strong model regularization in PReNet can lead to underfitting on those datasets. Therefore, a small $\lambda$, \eg, $\lambda=0.01$, is generally recommended.

\end{itemize}

\section{Conclusions}
This paper explores the generalized semi-supervised anomaly detection problem and introduces a novel deep approach and its instantiated model PReNet to address the problem. The approach learns pairwise relation features and anomaly scores in a unified framework by three-way pairwise relation modeling.
In doing so, it learns diverse abnormality and normality representations, alleviating the overfitting of the seen abnormalities. 
This is justified by the substantial improvement of PReNet over its variants and nine state-of-the-art competing methods that focus on learning more homogeneous normal/abnormal representations for detecting seen/unseen anomalies on 12 real-world datasets. 

Particularly, our significantly improved precision-recall performance in unseen anomaly detection, \ie, 10\%-30\%, is encouraging in that it is already very challenging to improve this metric for seen anomalies, and the challenge is further largely increased for the unseen anomalies.
Our results also suggest that the labeled anomaly data, regardless of its scale and coverage of the anomaly classes, can be well leveraged to enable accurate anomaly detection in the wild. In future work, we plan to extend our approach to explore the supervision information from different domains of data for anomaly detection.

\section*{Acknowledgments}
C. Shen's participation was supported by National Key R\&D Program of China (No. 2022ZD0118700). We thank Hezhe Qiao for helping obtain the results of XGBOD, PUMAD, and FEAWAD.

\appendix

\section{Proof of Theorem}\label{subsec:proof}

\begin{theorem}[Robustness to Anomaly Contamination]
Let $\epsilon\geq0$ be the anomaly contamination rate in $\mathcal{U}$, $y_{\mathbf{\{a,a\}}}=c_1$, $y_{\mathbf{\{a,u\}}}=c_2$, and $y_{\mathbf{\{u,u\}}}=c_3$ with $c_1>c_2>c_3\geq 0$, then for a given test instance $\mathbf{x}_k$, we have $\mathbb{E}\left[s_{\mathbf{x}_k} |\mathbf{x}_k\ is\ an\ anomaly \right] =\frac{c_1+c_2}{2}$, which is guaranteed to be greater than $\mathbb{E}\left[s_{\mathbf{x}_k} |\mathbf{x}_k\ is\ normal \right] =\frac{c_2+c_3+\epsilon (c_1+c_2)}{2}$ for $\epsilon <\frac{c_1-c_3}{c_1+c_2}$.
\end{theorem}

\begin{proof}
From the regression modeling, the expectation for the pairwise anomaly score for a true anomaly $\mathbf{a}_{k}$ is
\begin{align*}
\mathbb{E}\left[\phi\left(\left(\mathbf{a}_{i},\mathbf{a}_{k}\right);\Theta^{*}\right)\right]&= c_1,\\ 
\mathbb{E}\left[\phi\left(\left(\mathbf{a}_{k},\mathbf{u}_{j}\right);\Theta^{*}\right)\right]&= c_2, \\
\end{align*}

The corresponding expectation for a normal data instance $\mathbf{n}_{l}$ is given by
\begin{align*}
\mathbb{E}\left[\phi\left(\left(\mathbf{a}_{i},\mathbf{n}_{l}\right);\Theta^{*}\right)\right]&= 
\mathbb{E}\left[\phi\left(\left(\mathbf{a}_{i},\mathbf{u}_{j}\right);\Theta^{*}\right)\right] + \epsilon\mathbb{E}\left[\phi\left(\left(\mathbf{a}_{i},\mathbf{a}_{j}\right);\Theta^{*}\right)\right]\\
&=c_2+ \epsilon c_1,\\
\mathbb{E}\left[\phi\left(\left(\mathbf{n}_{l},\mathbf{u}_{j}\right);\Theta^{*}\right)\right]&= 
\mathbb{E}\left[\phi\left(\left(\mathbf{u}_{i},\mathbf{u}_{j}\right);\Theta^{*}\right)\right] + \epsilon\mathbb{E}\left[\phi\left(\left(\mathbf{a}_{i},\mathbf{u}_{j}\right);\Theta^{*}\right)\right]\\
&=c_3 + \epsilon c_2.\\
\end{align*}
Thus, based on the anomaly scoring function in Eqn. (\ref{eqn:score2}):
\begin{equation}\label{eqn:score}
    \mathit{s}_{\mathbf{x}_{k}} = \frac{1}{2E} \left[ \sum_{i=1}^{E} \phi\big(\left(\mathbf{a}_{i},\mathbf{x}_{k}\right);\Theta^{*}\big) + \sum_{j=1}^{E} \phi\big((\mathbf{x}_{k},\mathbf{u}_{j});\Theta^{*}\big)\right],
\end{equation}
we have 
\begin{multline}\label{eqn:anomalyscore}
\mathbb{E}\left[s_{\mathbf{x}_k} |\mathbf{x}_k\ is\ an\ anomaly \right] =\frac{1}{2E}\Big\{ E\times\mathbb{E}\left[\phi\left(\left(\mathbf{a}_{i},\mathbf{a}_{k}\right);\Theta^{*}\right)\right]\\+E\times\mathbb{E}\left[\phi\left(\left(\mathbf{a}_{k},\mathbf{u}_{j}\right);\Theta^{*}\right)\right]\Big\} =\frac{c_1+c_2}{2}, 
\end{multline}
and 
\begin{multline}\label{eqn:normalscore}
\mathbb{E}\left[s_{\mathbf{x}_k} |\mathbf{x}_k\ is\ normal \right] 
=\frac{1}{2E}\Big\{ E\times\mathbb{E}\left[\phi\left(\left(\mathbf{a}_{i},\mathbf{n}_{l}\right);\Theta^{*}\right)\right]\\+E\times\mathbb{E}\left[\phi\left(\left(\mathbf{n}_{l},\mathbf{u}_{j}\right);\Theta^{*}\right)\right]\Big\}
=\frac{c_2+c_3+\epsilon (c_1+c_2)}{2}.
\end{multline}

Since $c_1>c_2>c_3\geq 0$, $\mathbb{E}\left[s_{\mathbf{x}_k} |\mathbf{x}_k\ is\ an\ anomaly \right]$ is guaranteed to be greater than $\mathbb{E}\left[s_{\mathbf{x}_k} |\mathbf{x}_k\ is\ normal \right]$ for $\epsilon <\frac{c_1-c_3}{c_1+c_2}$.
\end{proof}

\section{Detailed Experiment Setup}

\subsection{Datasets}\label{subsec:data}

\subsubsection{Seen Anomaly Detection Datasets} The 12 seen anomaly detection datasets can be accessed via the links in Table \ref{tab:link}. More specifically, the \textit{donors} data is taken from KDD Cup 2014 for predicting the excitement of donation projects, with exceptionally exciting projects used as anomalies (6.0\% of all data instances). The \textit{census} data is extracted from the US census bureau database, in which we aim to detect the rare high-income persons (6.0\%). \textit{fraud} is for fraudulent credit card transaction detection, with fraudulent transactions (0.2\%) as anomalies. \textit{celeba} contains more than 200K celebrity images, each with 40 attribute annotations. We use the bald attribute as our detection target, in which the scarce bald celebrities (3.0\%) are treated as anomalies and the other 39 attributes form the feature space. The \textit{dos}, \textit{rec}, \textit{fuz} and \textit{bac} datasets are derived from a popular intrusion detection dataset called \textit{UNSW-NB15} \cite{moustafa2015nb15} with the respective \textit{DoS} (15.0\%), \textit{reconnaissance} (13.1\%), \textit{fuzzers} (3.1\%) and \textit{backdoor} (2.4\%) attacks as anomalies against the `normal' class. \textit{w7a} is a web page classification dataset, with the minority classes (3.0\%) as anomalies. \textit{campaign} is a dataset of bank marketing campaigns, with rare positive campaigning records (11.3\%) as anomalies. \textit{news20} is one of the most popular text classification corpora, which is converted into anomaly detection data via random downsampling of the minority class (5.0\%) based on \cite{liu2012iforest,zimek2013subsampling}. \textit{thyroid} is for disease detection, in which the anomalies are the hypothyroid patients (7.4\%). Seven of these datasets contain real anomalies,including \textit{donors}, \textit{fraud}, \textit{dos}, \textit{rec}, \textit{fuz}, \textit{bac} and \textit{thyroid}. The other five datasets contain semantically real anomalies, \ie, they are rare and very different from the majority of data instances. So, they serve as a good testbed for the evaluation of anomaly detection techniques.

\begin{table}[htbp]
    \centering
\caption{Links for accessing the datasets}
\scalebox{0.78}{
    \begin{tabular}{l|p{8.0cm}}
    \hline
    \textbf{Data} & \textbf{Link}\\\hline
    donors &https://www.kaggle.com/c/kdd-cup-2014-predicting-excitement-at-donors-choose \\
    census & https://archive.ics.uci.edu/ml/datasets/census+income\\
    fraud&https://www.kaggle.com/c/1056lab-credit-card-fraud-detection \\
    celeba&http://mmlab.ie.cuhk.edu.hk/projects/CelebA.html\\
    dos & https://www.unsw.adfa.edu.au/unsw-canberra-cyber/cybersecurity/ADFA-NB15-Datasets/\\
    rec&https://www.unsw.adfa.edu.au/unsw-canberra-cyber/cybersecurity/ADFA-NB15-Datasets/\\
    fuz&https://www.unsw.adfa.edu.au/unsw-canberra-cyber/cybersecurity/ADFA-NB15-Datasets/\\
    bac&https://www.unsw.adfa.edu.au/unsw-canberra-cyber/cybersecurity/ADFA-NB15-Datasets/\\
    w7a&https://www.csie.ntu.edu.tw/$\sim$cjlin/libsvmtools/datasets/\\ 
    campaign&https://archive.ics.uci.edu/ml/datasets/bank+marketing\\
    news20&https://www.csie.ntu.edu.tw/$\sim$cjlin/libsvmtools/datasets/\\
    thyroid&https://www.openml.org/d/40497 \\
    \hline
    \end{tabular}
}
\label{tab:link}
\end{table}

To replicate the real-world scenarios where we have a few labeled anomalies and large unlabeled data, we first have a stratified split of the anomalies and normal instances into two subsets, with 80\% data as training data and the other 20\% data as a holdup test set. Since the unlabeled data is often anomaly-contaminated, we then combine some randomly selected anomalies with the whole normal training data instances to form the unlabeled dataset $\mathcal{U}$. We further randomly sample a limited number of anomalies from the anomaly class to form the labeled anomaly set $\mathcal{A}$. The resulting sample size and dimensionality of the datasets are shown in Table \ref{tab:rocpr2}.

\subsubsection{Unseen Anomaly Detection Datasets}
Table \ref{tab:unseen} presents the 28 datasets for the evaluation of detecting unseen anomalies. These datasets are derived from the above four intrusion attack datasets \textit{dos}, \textit{rec}, \textit{fuz} and \textit{bac}, with data instances spanned by the same feature space. To guarantee that the evaluation data contains unseen anomalies, the anomaly class in one of these four datasets is held up for evaluation, while the anomalies in any combinations of the remaining three datasets are combined to form the pool of seen anomalies. The type of the holdup anomalies is always different from that in the anomaly pool and can be safely treated as unseen anomalies. We have 28 possible permutations under this setting, resulting in 28 datasets with different seen and/or unseen anomalies. For the training, $\mathcal{A}$ contains the anomalies sampled from the seen anomalies pool, while the evaluation data is composed of the holdup unseen anomaly class and the 20\% holdup normal instances. Note that the other eight datasets in Table \ref{tab:rocpr2} cannot be used in evaluating unseen anomaly detection as they contain only one anomaly class and they are from different data sources and feature spaces.

\subsection{Implementation Details}

\subsubsection{Packages}

PReNet is implemented using Tensorflow/Keras. The main packages and their versions used in this work are provided as follows:
\begin{itemize}
    \item keras==2.3.1
    \item numpy==1.16.2
    \item pandas==0.23.4
    \item scikit-learn==0.20.0
    \item scipy==1.1.0
    \item tensorboard==1.14.0
    \item tensorflow==1.14.0
\end{itemize}

\subsubsection{Hyperparameter Settings}\label{sec:competing2}

Since our experiments focus on unordered multidimensional data, multilayer perceptron networks are used. Similar to \cite{pang2018repen,pang2019devnet}, we empirically found that all deep methods using an architecture with one hidden layer perform better and more stably than using two or more hidden layers. This may be due to the limit of the available labeled data. Following \cite{pang2018repen,pang2019devnet}, one hidden layer with 20 neural units is used in PReNet. The ReLu activation function $g(a) = \mathit{max}(0, a)$ is used. An $\ell_2$-norm regularizer with the hyperparameter setting $\lambda=0.01$ is applied to avoid overfitting. The RMSprop optimizer with the learning rate $0.001$ is used. The same network architecture is used in the competing methods DevNet \cite{pang2019devnet}, REPEN \cite{pang2018repen}, Deep SAD (DSAD) \cite{ruff2019deep}, FSNet \cite{snell2017protonet} and its variant cFSNet. iForest with the recommended settings \cite{liu2012iforest}, 100 isolation trees and 256 subsampling size, are used in our experiments.

All deep detectors are trained using 50 epochs, with 20 batches per epoch. The batch size is probed in $\{$8, 16, 32, 64, 128, 256, 512$\}$. The best fits, 512 in PReNet, DevNet and DSAD, 256 in FSNet and REPEN, are used by default. cFSNet uses the same settings as FSNet. Similar to PReNet, oversampling is also applied to the labeled anomaly set $\mathcal{A}$ to well train the deep detection models of DevNet, REPEN, DSAD, FSNet and cFSNet. 

\section{Additional Empirical Results}\label{sec:additional_results}

\subsection{Comparison to Three More Methods}

We also compare PReNet to three more methods XGBOD \cite{zhao2018xgbod}, PUMAD \cite{ju2020pumad}, and FEAWAD \cite{zhou2022feature} on both seen and unseen anomaly detection datasets, with the results shown in Tables \ref{tab:new_seenad} and \ref{tab:new_unseenad} respectively. The official implementation of XGBOD and FEAWAD was used to perform the experiments. The code of PUMAD is not released; we use our own implementation based on a metric learning similar to REPEN. As for XGBOD, we used one-class SVM and iForest to produce new features only; $k$NN and LOF were excluded due to prohibitive computational cost on large datasets. All the other settings in XGBOD remain unchanged.

\begin{table}[!htbp]
  \centering
  \caption{AUC-PR results for seen AD datasets. OOM denotes an out-of-memory issue in a GeForce RTX 3090 24GB GPU.}
  \scalebox{0.85}{
    \begin{tabular}{lcccc}
    \hline  
    \textbf{Dataset} & \textbf{PReNet} & \textbf{XGBOD} & \textbf{PUMAD} & \textbf{FEAWAD} \\\hline
    donors  & \textbf{1.000} & 0.178 & 0.215 & \textbf{1.000} \\
    census  & \textbf{0.356} & 0.061 & 0.129 & 0.252 \\
    fraud  & \textbf{0.689} & 0.408 & 0.423 & 0.670 \\
    celeba  & \textbf{0.309} & 0.081 & 0.143 & 0.225 \\
    dos    & \textbf{0.900} & 0.429 & 0.328 & 0.827 \\
    rec    & 0.767 & 0.041 & 0.379 & \textbf{0.852} \\
    fuz    & \textbf{0.170} & 0.097 & 0.065 & 0.118 \\
    bac    & \textbf{0.890} & 0.145 & 0.120 & 0.811 \\
    w7a    & \textbf{0.496} & 0.224 & 0.072 & 0.406 \\
    campaign  & \textbf{0.470} & 0.323 & 0.302 & 0.365 \\
    news20  & \textbf{0.652} & 0.076 & 0.107 & OOM \\
    thyroid  & 0.298 & 0.262 & 0.163 & \textbf{0.383} \\\hline
    \end{tabular}%
    }
  \label{tab:new_seenad}%
\end{table}%

The results show that these three methods, especially FEAWAD, can work well on some datasets, but they still substantially underperform our method PReNet on most datasets. 

\begin{table}[!hbp]
  \centering
  \caption{AUC-PR results for unseen AD datasets.}
  \scalebox{0.85}{
    \begin{tabular}{llcccc}
    \hline
    \textbf{Seen} & \textbf{Unseen} & \textbf{PReNet} & \textbf{XGBOD} & \textbf{PUMAD} & \textbf{FEAWAD} \\\hline
    dos     & bac   & \textbf{0.908} & 0.429 & 0.360 & 0.832 \\
    dos, fuz   & bac   & \textbf{0.889} & 0.372 & 0.274 & 0.703 \\
    fuz   & bac   & \textbf{0.503} & 0.145 & 0.061 & 0.131 \\
    rec   & bac   & 0.752 & 0.224 & 0.497 & \textbf{0.851} \\
    rec, dos   & bac   & \textbf{0.834} & 0.367 & 0.321 & 0.831 \\
    rec, dos, fuz   & bac   & \textbf{0.706} & 0.378 & 0.143 & 0.610 \\
    rec, fuz   & bac   & \textbf{0.711} & 0.284 & 0.138 & 0.533 \\
    bac   & dos   & \textbf{0.938} & 0.097 & 0.114 & 0.785 \\
    bac, fuz   & dos   & \textbf{0.932} & 0.284 & 0.053 & 0.176 \\
    fuz   & dos   & \textbf{0.811} & 0.145 & 0.061 & 0.128 \\
    rec   & dos   & \textbf{0.928} & 0.224 & 0.388 & 0.848 \\
    rec, bac   & dos   & \textbf{0.891} & 0.244 & 0.164 & 0.772 \\
    rec, bac, fuz   & dos   & \textbf{0.835} & 0.178 & 0.199 & 0.550 \\
    rec, fuz   & dos   & \textbf{0.883} & 0.228 & 0.136 & 0.556 \\
    bac   & fuz   & 0.418 & 0.097 & 0.114 & \textbf{0.769} \\
    dos     & fuz   & 0.418 & 0.429 & 0.328 & \textbf{0.842} \\
    dos, bac   & fuz   & 0.375 & 0.284 & 0.189 & \textbf{0.805} \\
    rec   & fuz   & 0.462 & 0.224 & 0.388 & \textbf{0.860} \\
    rec, bac   & fuz   & 0.315 & 0.244 & 0.166 & \textbf{0.790} \\
    rec, bac, dos   & fuz   & 0.294 & 0.342 & 0.241 & \textbf{0.734} \\
    rec, dos   & fuz   & 0.349 & 0.361 & 0.321 & \textbf{0.840} \\
    bac   & rec   & \textbf{0.892} & 0.097 & 0.155 & 0.684 \\
    bac, fuz   & rec   & \textbf{0.876} & 0.054 & 0.042 & 0.192 \\
    dos     & rec   & \textbf{0.849} & 0.429 & 0.335 & 0.818 \\
    dos, bac   & rec   & 0.768 & 0.284 & 0.163 & \textbf{0.775} \\
    dos, bac, fuz   & rec   & \textbf{0.719} & 0.163 & 0.153 & 0.700 \\
    dos, fuz   & rec   & \textbf{0.788} & 0.371 & 0.280 & 0.700 \\
    fuz   & rec   & \textbf{0.797} & 0.145 & 0.071 & 0.745 \\\hline
    \end{tabular}%
    }
  \label{tab:new_unseenad}%
\end{table}%

\subsection{Performance Ranking of All Methods}

To have a holistic comparison of all 10 detectors, we calculate the average (ordinal and percentile) rank of each method based on its detection performance in both seen and unseen AD settings. The results are shown in Table \ref{tab:avg_rank}, where an average ordinal rank of one (or a percentile of 1.00 ) is the perfect performance, indicating the method is always the best performer compared to all other methods across all datasets. That is, lower rank (or higher percentile) indicates better performance.

The results show that PReNet is the best performer in both settings, outperforming all nine competing methods for 94.2\% and 90.2\% of cases on 12 seen anomaly detection datasets and 28 unseen anomaly detection datasets respectively. It is followed by DevNet and DSAD in seen AD, and DevNet and FEAWAD in unseen AD. When using a Conover post-hoc test at the 95\% confidence level, PReNet performs significant better than all methods except DevNet and DSAD in seen AD; it significantly outperforms all other nine methods in unseen AD.

\begin{table}[!htbp]
  \centering
  \caption{Average (ordinal and percentile) rank of methods based on AUC-PR for seen and unseen AD across the 12 and 28 datasets, respectively. The methods are sequentially sorted based on the ordinal rank of seen and unseen AD.}
  \scalebox{0.85}{
    \begin{tabular}{lcc|cc}
    \hline
          & \multicolumn{2}{c|}{\textbf{Seen AD}} & \multicolumn{2}{c}{\textbf{Unseen AD}} \\\hline
    \textbf{Method} & \textbf{Ordinal} & \textbf{Percentile} & \textbf{Ordinal} & \textbf{Percentile} \\\hline
    PReNet & \textbf{1.583} & \textbf{0.942} & \textbf{1.982} & \textbf{0.902} \\
    DevNet & 2.667 & 0.832 & 3.661 & 0.734 \\
    DSAD  & 3.292 & 0.769 & 4.143 & 0.686 \\
    FEAWAD & 3.955 & 0.705 & 3.786 & 0.721 \\
    cFSNet & 5.792 & 0.517 & 4.518 & 0.648 \\
    REPEN & 6.583 & 0.439 & 4.893 & 0.611 \\
    FSNet & 6.625 & 0.432 & 6.054 & 0.495 \\
    XGBOD & 7.417 & 0.352 & 7.857 & 0.314 \\
    PUMAD & 7.583 & 0.337 & 8.750 & 0.225 \\
    iForest & 9.000 & 0.193 & 9.357 & 0.164 \\\hline
    \end{tabular}%
    }
  \label{tab:avg_rank}%
\end{table}%

\bibliographystyle{ACM-Reference-Format}
\balance
\bibliography{references}

\end{document}